%% file: emnlp2025.tex
\pdfoutput=1

\documentclass[11pt]{article}

\usepackage[final]{acl}

\usepackage{times}
\usepackage{latexsym}
\usepackage{float}
\usepackage{graphicx}
\usepackage{amsmath}

\usepackage[T1]{fontenc}

\usepackage[utf8]{inputenc}

\usepackage{microtype}
\usepackage{xcolor}         
\usepackage{makecell}
\usepackage{wrapfig,lipsum}
\usepackage{bbold}
\usepackage{url}
\usepackage{hyperref}
\usepackage{CJKutf8}
\usepackage{graphicx}
\usepackage{times}
\usepackage{pdfpages}
\usepackage{amsmath}
\usepackage{tablefootnote}
\usepackage{tcolorbox}
\usepackage{colortbl}
\usepackage{float}
\usepackage{multirow}
\usepackage{booktabs}
\usepackage{latexsym}
\usepackage{listings}
\usepackage{bbding}
\usepackage{pifont}
\usepackage{wasysym}
\usepackage{amssymb}
\usepackage{paralist}
\usepackage{spverbatim}
\usepackage{fancyvrb}
\usepackage{fvextra}
\usepackage{adjustbox}
\usepackage[table,dvipsnames]{xcolor}

\definecolor{Blueback}{RGB}{218, 227, 243} 
\definecolor{Greenback}{RGB}{226, 240, 217}
\definecolor{Redback}{RGB}{251, 229, 214} 
\definecolor{shadecolor}{rgb}{0.5,0.5,0.5}
\definecolor{pastelgray}{rgb}{0.81, 0.81, 0.77}
\definecolor{snow}{rgb}{1, 0.98, 0.98}
\definecolor{papayawhip}{rgb}{1.0, 0.94, 0.84}
\definecolor{c1}{HTML}{9C4A1A}
\definecolor{airforceblue}{rgb}{0.796,0.878,0.937}
\definecolor{airforceblue}{rgb}{0.828,0.914,0.910}
\definecolor{lightblue}{rgb}{0.933,0.968,0.988}
\definecolor{codeblue}{rgb}{0.215,0.686,0.847}
\definecolor{ora}{rgb}{0.914,0.443,0.196}

\newcommand\modelname{\textsc{MM-Critic}}

\title{\modelname: A Holistic Evaluation of Large Multimodal Models as Multimodal Critique}

\author{%
  $^{\spadesuit}$~$^{\diamondsuit}$Gailun Zeng~,
  $^{\spadesuit}$Ziyang Luo~,
  $^{\spadesuit}$Hongzhan Lin~,
  $^{\spadesuit}$Yuchen Tian~\\
  $^{\clubsuit}$\textbf{Kaixin Li}~,
  $^{\heartsuit}$\textbf{Ziyang Gong}~,
  $ ^{\bigstar}$~$^{\diamondsuit}$\textbf{Jianxiong Guo}\footnotemark[1]~,
  $^{\spadesuit}$\textbf{Jing Ma}\thanks{~~Corresponding Authors.}\\[1ex]
  $^\spadesuit$Hong Kong Baptist University, $^\diamondsuit$Beijing Normal-Hong Kong Baptist University\\
  $^\clubsuit$National University of Singapore, $^\bigstar$Beijing Normal University \\
  $^\heartsuit$Shanghai Jiao Tong University \\
  \texttt{gailun\_zeng@u.nus.edu},
  \texttt{jianxiongguo@bnu.edu.cn} \\
  \texttt{majing@comp.hkbu.edu.hk}  
}

\begin{document}
\maketitle
\begin{abstract}

The ability of critique is vital for models to self-improve and serve as reliable AI assistants. While extensively studied in language-only settings, multimodal critique of Large Multimodal Models (LMMs) remains underexplored despite their growing capabilities in tasks like captioning and visual reasoning.
In this work, we introduce \modelname{}, a holistic benchmark for evaluating the critique ability of LMMs across multiple dimensions: basic, correction, and comparison. Covering 8 main task types and over 500 tasks, \modelname{} collects responses from various LMMs with different model sizes and is composed of 4471 samples.
To enhance the evaluation reliability, we integrate expert-informed ground answers into scoring rubrics that guide GPT-4o in annotating responses and generating reference critiques, which serve as anchors for trustworthy judgments. Extensive experiments validate the effectiveness of \modelname{} and provide a comprehensive assessment of leading LMMs’ critique capabilities under multiple dimensions. Further analysis reveals some key insights, including the correlation between response quality and critique, and varying critique difficulty across evaluation dimensions. Our code is available at \href{https://github.com/MichealZeng0420/MM-Critic}{https://github.com/MichealZeng0420/MM-Critic}.

\end{abstract}

\input{sections/1.Intro}
\input{sections/2.Related_work}
\input{sections/4.MMCritic}

\input{sections/5.Experiments}
\input{sections/6.Conclusion}

\section*{Limitations}
Note that there are still some drawbacks and limitations about \modelname{}.
\begin{itemize}
    \item Although \modelname{} is a comprehensive benchmark, it currently focuses only on text and image modalities, lacking broader evaluation across other multimodal domains such as video, audio, and 3D data.
    \item The reference critiques and scoring annotations are generated by GPT-4o guided by rubric checklists. While this design improves consistency, it still relies on a single model as the annotator, which may introduce systematic biases or limitations inherent to GPT-4o.  
    \item \modelname{} evaluates model critique in a static context using predefined prompts and samples. In real-world scenarios, critique often occurs interactively or iteratively, which is not yet captured by the current benchmark.  
\end{itemize}

\section*{Acknowledgements}
This work is partially supported by Tencent Rhino-Bird Focused Research Program (Value-aligned Credible Large Language Model), the National Natural Science Foundation of China (NSFC) under Grant No.~62202055, the Guangdong Basic and Applied Basic Research Foundation under Grant No.~2025A1515012843, the Start-up Fund from Beijing Normal University under Grant No.~312200502510, the Internal Fund from Beijing Normal-Hong Kong Baptist University under Grant No.~UICR0400003-24  and No.~UICR0200022-25, and the Interdisciplinary Intelligence SuperComputer Center of Beijing Normal University (Zhuhai).

\bibliography{anthology,custom}
\bibliographystyle{acl_natbib}


\input{sections/7.appendix}

\end{document}

%% file: sections/1.Intro.tex
\section{Introduction}
\label{set:intro}


The critique ability of language models plays a pivotal role in fostering self-improvement~\citep{liu2024divingselfevolvingtrainingmultimodal} and enabling trustworthy AI~\cite{DBLP:journals/corr/abs-2311-02801, lin-etal-2025-fact}, e.g., critique-capable models can provide feedback on student answers or essays, supporting personalized learning in educational applications~\cite{Parker_2024}. This capability has been extensively explored in the context of Large Language Models (LLMs)~\citep{lan2024criticeval,CriticBench,DBLP:conf/acl/SunLYYLL24}. However, as Large Multimodal Models (LMMs) gain proficiency across diverse multimodal tasks involving captioning~\citep{AuroraCap} and visual reasoning~\citep{MMReasoning}, their potential to analyze and critique becomes increasingly important, not only for refining their own outputs but also for serving as AI assistants capable of providing feedback in complex, real-world scenarios~\citep{LLaVA-Critic,luo2025mcp}. The rise of LMMs brings new challenges and opportunities for critique in multimodal contexts. For example, LMMs must reason over and align information from multiple modalities (e.g., image and text), which introduces complexity in both understanding and critique generation. Thus, evaluating such critique capabilities in LMMs is critical for advancing their alignment, reliability, and reasoning depth across modalities~\citep{MMRewardBench,VLRewardBench}.

\input{figure/figure_open}
\input{tables/Table_Compare}

Recently, several efforts have been made to evaluate or enhance the critique capabilities of LMMs. Notably, Multimodal RewardBench~\citep{MMRewardBench} and VL-RewardBench~\citep{VLRewardBench} investigate the judging abilities of LMMs by presenting two responses to a multimodal question and asking the model to select the better one. These benchmarks primarily frame the critique as a binary classification task focused on simple preference prediction, without delving into more fine-grained aspects of the critique capabilities. Beyond preference modeling, LLaVA-Critic~\citep{LLaVA-Critic} introduces an open-source LMM trained to effectively evaluate the responses of other LMMs. While it showcases the potential of LMMs for judging, it is primarily designed for model training rather than systematic evaluation. Similarly, Critic-V~\citep{Critic-V} explores the use of LMMs as critics to catch errors made in multimodal reasoning tasks. Although it demonstrates that LMMs can act as effective critics, it focuses on case studies and empirical validation, rather than establishing a comprehensive benchmark for critique capability. Taken together, these works highlight the growing interest in multimodal critique, yet reveal a lack of standardized, holistic evaluation that assesses LMMs as general-purpose multimodal critics across tasks and critique dimensions.

To fill this gap, we propose a novel benchmark, \modelname{}, designed to comprehensively and reliably measure critique capability of LMMs. To ensure the comprehensiveness of \modelname{}, firstly, we propose a granular evaluation scheme, where we employ both scalar and textual metrics to evaluate the critique capabilities of LMMs across three dimensions, namely\textit{ basic critique, correction critique, and comparative critique}, as shown in Figure~\ref{fig:motivation}.
Second, \modelname{} sources diverse data from MEGA-BENCH~\cite{chen2024mega}, a comprehensive multimodal task benchmark encompassing 8 main task scenarios and over 500 specific tasks. Then, we leverage a broad range of LMMs with different model sizes to collect enough responses based on the selected specific tasks, which ensures that the generated responses exhibit distinguishable levels of quality. Finally, \modelname{} totally includes 4471 model response samples. Based on this,  we also organized sub-datasets for fine-grained critique evaluation, namely correction critique and comparative critique.

To enhance the reliability of evaluation, we incorporate reference critiques to assist the judge model~(i.e., GPT-4.1) in evaluating LMMs' critiques. This effectively mitigates potential evaluation bias in textual critique introduced by judge models (e.g., GPT-4.1)~\cite{wang2023shepherd,li2024generation,tan2024large}. Specifically, considering the characteristics of task types, we deliberately design a detailed scoring rubric checklist that include both common rubrics and task-type-specific rubrics. Besides, we also provide each task's grounded answer to the annotation model~(i.e., GPT-4o) since these selected tasks belong to different domains and the grounded answer, including expert-human level knowledge, helps the annotator generate both reasonable and reliable reference critiques. Then, the checklist combined with grounded answers is embedded into the prompt to guide GPT-4o in annotating both the response quality scores and reference critiques. Based on the annotated response quality, it is feasible to construct the sub-dataset for correction critique from low-quality responses and generate the sub-dataset for comparative critique by forming response pairs with different quality levels.  Overall, as shown in Table~\ref{tab:my_bench}, \modelname{} demonstrates substantial improvements in terms of comprehensiveness over prior benchmarks.

The reference critiques help us define \textit{Critique Score} metrics that can score the textual critique contents generated by LMMs, where we anchor the reference critiques at a score of 8 to represent human levels and prompt the judge model to compare LMMs' contents with the reference critiques and give comparative scores. We also employ the common \textit{Critique/Preference Accuracy} as scalar metrics.
We conduct extensive experiments on leading closed-source and open-source LMMs. The results validate the effectiveness of \modelname{} and reveal LMMs' critique capabilities, where the scaling law is clearly observed and models within the same series exhibit consistently improved critique performance as their parameter sizes increase. Extensive case analysis demonstrates that this approach of reference-critique-based evaluation significantly enhances the reliability of the judgments. 
 Then, further experiments and analyses reveal a set of implicit yet intriguing insights:
\begin{itemize}
\item Correction critique scores are generally lower than basic critique scores, indicating that Correction critique remains a challenging task for LMMs. In comparative critique, pairwise combinations of medium/high-quality responses are particularly difficult to judge.
\item There exists an inherent relationship between response quality and critique scores. Results show that medium-quality responses tend to receive the lowest critique scores compared to both high- and low-quality ones, highlighting the unique challenges posed by evaluating critiques of medium-quality responses.
\item The judgment bias of models may be related to the richness of critique text, as GPT-4.1 tends to assign higher scores to longer, more elaborate critiques.
\end{itemize}

%% file: figure/figure_open.tex
\begin{figure*}[h]
    \centering
    \includegraphics[width=\linewidth]{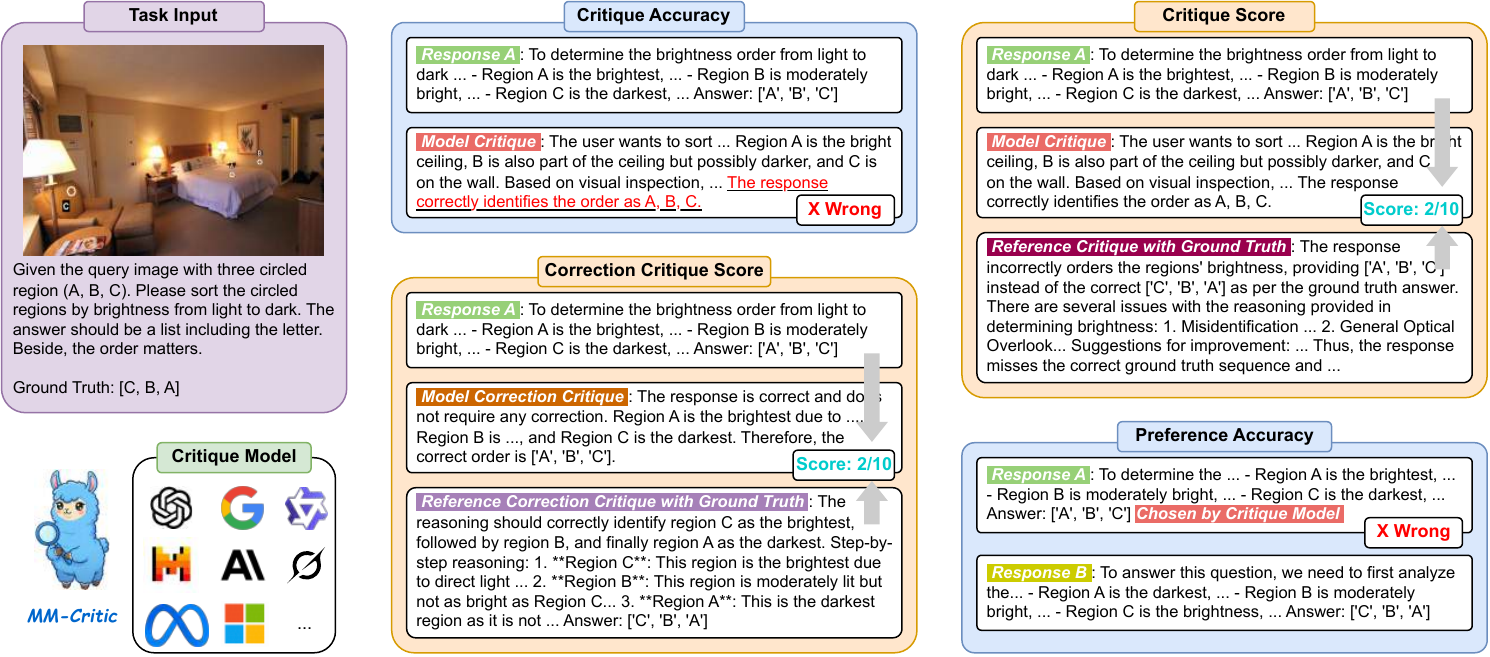}
    \caption{Multi-dimensional critique evaluation in \modelname{}. Basic critique includes binary correctness and textual feedback (\textit{Critique Accuracy}, \textit{Critique Score}); correction and comparative critique correspond to \textit{Correction Critique Score} and \textit{Preference Accuracy}, respectively.}
    \label{fig:motivation}
\end{figure*}

%% file: tables/Table_Compare.tex

\begin{table*}[t]
  \small
  \centering
  \resizebox{\textwidth}{!}{%
    \begin{tabular}{lcccc}
      \toprule
      Benchmarks & Critique Metric & Critique Dimension & Task Types & Taxonomy Hierachy \\
      \midrule
      MLLM-as-a-Judge~\cite{MLLM-as-a-Judge} & Scalar & 3 & 12 & 1 \\
      Multimodel RewardBench~\cite{MMRewardBench} & Scalar & 1 & 6 & 1 \\
      VL-RewardBench~\cite{VLRewardBench} & Scalar & 1 & 3 & 1 \\
      \midrule
      \modelname{} (ours) & Scalar/Textual & 3 & 8 & 3 \\
      \bottomrule
    \end{tabular}%
  }
  \caption{Comparison between related benchmarks and \modelname{}.}
  \label{tab:my_bench}
\end{table*}

%% file: sections/2.Related_work.tex
\section{Related work}
\label{set: RW}

\paragraph{Application.} The critique ability of models has been extensively explored in recent years as a means of assessing response quality across a variety of tasks while reducing reliance on costly human annotations~\cite{DBLP:journals/corr/abs-2308-03188,DBLP:journals/corr/abs-2308-04592,MTBench}. Advanced LLMs, such as GPT-4, have demonstrated strong alignment with human judgments~\cite{JudgeBench}, prompting the extension of this paradigm to multimodal settings. SOTA LMMs, including GPT-4o, are increasingly employed to evaluate responses in multimodal tasks, significantly alleviating the need for manual evaluation~\cite{VideoAutoArena}.

Beyond judging, critique also plays a crucial role in enhancing LMM performance. During inference, textual critiques that identify response flaws and suggest improvements enable iterative refinement~\cite{Self-Refine}. During training, scalar-valued critique signals are commonly used to construct response pairs with clear quality differences~\cite{liu2024divingselfevolvingtrainingmultimodal}, facilitating methods such as preference learning to further boost model capabilities~\cite{DBLP:journals/corr/abs-2402-10884}. Besides, critique capability facilitates a series of downstream applications, such as harmful content detection~\cite{chen-etal-2025-adammeme, lin2024towards,huang2024towards}, sarcasm understanding~\cite{chen2024cofipara} and GUI-based tasks~\cite{yang2025gta1,li2025screenspot}. 

\paragraph{Evaluation.} 
Due to multimodal complexity, it is non-trivial to reasonably evaluate LLMs' performance on specific applications, necessitating the importance of task-specific benchmarkings. Recently, diverse evaluation frameworks have emerged.  Multimodal trustworthy AI shows new challenges in fact checking~\cite{wang2024mfc} and harmful content audit~\cite{lin2024goat,lin2023beneath}. Coding, as the widely-discussed research direction, when considering rich visual programming environments, the evaluation and exploration of LLMs' capabilities is valuable ~\cite{li2024mmcode,fu2024scratcheval}. Deep understanding of visual components is still unexplored~\cite{gong2025space,yang2024aria}. These benchmarks are conducive to better master models' deficiency and carry out targeted model capability enhancement~\cite{cheng2024videollama}.

A range of benchmarks has been developed to assess the critique capabilities of models. Initial efforts predominantly focused on the language domain, evaluating models’ ability to judge text-based responses~\cite{CriticBench,lan2024criticeval}. More recent work has extended this evaluation to the multimodal setting, primarily using scalar-valued critiques to measure alignment with human judgments on standard multimodal tasks~\cite{VLRewardBench,MMRewardBench,MLLM-as-a-Judge}. As shown in Table 1, our proposed MM-Critic advances beyond existing benchmarks by incorporating richer critique dimensions and a wider variety of task types, enabling a more comprehensive and nuanced evaluation of critique ability.

%% file: sections/4.MMCritic.tex
\section{\modelname{} Construction}
\label{set:benchmark}

\subsection{Overview}
\modelname{} is a holistic evaluation benchmark for multimodal tasks, covering 8 major task categories and over 500 specific tasks. An overview of \modelname{} is presented in Table~\ref{tab:data-response-collect} (Appendix~\S\ref{Apx:A}), comprising 4471 samples distributed across four sub-datasets: core, core single-image, open, and open single-image.

The core and core single-image sub-datasets include large-scale and commonly seen tasks, formulated as closed-ended questions with unique ground-truth answers. In contrast, the other two sub-datasets contain open-ended questions, for which only reference answers are provided.
Each sample also contains a question and a response generated by various LMMs, along with a series of annotations, such as response quality scores and reference textual critiques, produced by GPT-4o based on a rigorous scoring rubric checklist. These annotations play a crucial role in enhancing the evaluation process, where the reference critique can significantly improve the reliability of the judge model’s assessments.

The construction of \modelname{} follows a three-step process:
1)~selecting diverse samples from a comprehensive multimodal benchmark~\cite{chen2024mega} and using a series of LMMs to generate a wide range of responses;
2)~designing a rigorous scoring rubric checklist to guide GPT-4o in evaluating response quality and generating reference critiques in a human-expert-like manner;
3)~constructing sub-datasets for correlation and comparative critique based on the annotated response quality.

\subsection{Multimodal Task and Response Collection}
\label{mtrc}
\modelname{} is constructed based on MEGA-BENCH \cite{chen2024mega}, a comprehensive evaluation suite encompassing over 500 real-world multimodal tasks across 8 distinct categories.
To build the original dataset (see Table~\ref{tab:data-colllect} in Appendix~\S\ref{Apx:A}), we first randomly sampled two instances from each specific task, covering a wide spectrum of mainstream text-image tasks, thereby ensuring the diversity and representativeness of \modelname{}.
Subsequently, we employed a range of LMMs with varying capability levels (see Table~\ref{tab:llms_list} in Appendix~\S\ref{Apx:A}) to generate responses at different quality levels.
Through this dual-faceted approach, i.e., diverse task coverage and stratified response generation, \modelname{} serves as a holistic benchmark for evaluating LMMs' critique capability.

\subsection{Reference Critique Construction}
 Notice that the reliability of model-based judging methods remains an open question, and the bias between human and model-based evaluations still poses a significant challenge \cite{li2024generation,tan2024large}. To partially mitigate this bias and enhance the reliability of model judges (e.g., GPT-4o), we designed a detailed scoring rubric checklist and employed it with each task’s grounded answer to guide GPT-4o in generating reasonable annotations. Note that grounded answers include rich human-expert knowledge since \modelname{} is composed of different domains, such as coding and mathematics, and truly needs domain-specific experts to provide professional answers.
The complete scoring rubric checklist and prompt can be found in Table~\ref{tab:rubric} and Figure~\ref{ap:prompt_level1} (Appendix~\S\ref{Apx:B}). The following lists all annotations:

 \textbf{Correctness.}~We utilize the GPT-4o to judge the correctness of the generated responses by LMMs, where the task answers are available.

 \textbf{Response Quality Score.}~GPT-4o assigns each response a score ranging from $0$ to $10$ based on the rigorous scoring rubric checklist. This scoring process aims to stratify response quality and assist further 
 analysis, such as revealing correlations between response quality and the generated critiques.

 \textbf{Reference Critique.}~In addition to scalar evaluation, textual analysis is more complex yet essential, as the textual content provides deeper insight into each LMM’s critique capability. Guided by the scoring rubric checklist and the given human-expert answer, GPT-4o is prompted to emulate human experts in generating a textual reference critique. This reference critique is considered high-quality and anchored at a score of 8.
 
 \textbf{Reference Correction Critique.} For relatively low/medium-quality responses, it is reasonable to generate correction critiques that reflect the self-improvement potential of LMMs. Therefore, for suboptimal responses, the correction critiques generated by GPT-4o with access to the ground-truth answers, can be regarded as reliable and convincing feedback.

 \subsection{Sub-datasets for Correction and Comparative Critique}
We constructed dedicated sub-datasets tailored to the two critique dimensions. For correctness critique, we derived a subset from the core dataset of \modelname{}, deliberately selecting samples labeled with low or medium response quality. For comparative critique, we construct three pairwise sub-datasets from the core dataset of \modelname{}, leveraging the labeled response quality scores. Specifically, responses with scores in the range of $[0,4]$ are categorized as low-quality, $[5,7]$ as medium-quality, and $[8,10]$ as high-quality. Based on this categorization, we generate three types of pairwise combinations: (low, medium), (medium, high), and (low, high). Table~\ref {tab:data-subset-corec-compa} (Appendix~\S\ref{Apx:A}) lists the detailed statistics of the sub-datasets.

\input{tables/Table_Main_Results}

\section{Evaluation Metric}
\subsection{Comprehensive Critique Dimensions}
It is essential to consider evaluation metrics comprehensively and especially ensure that they align with our scenarios, namely LMM's critique. Following previous work \cite{lan2024criticeval,zhang2025codecriticbench}, \modelname{} is designed to thoroughly evaluate the critique abilities of LMMs across multiple dimensions. From the perspective of quantifiability, evaluation metrics can be categorized into scalar and textual forms. To ensure a rigorous assessment, we adopt a suite of metrics covering both \texttt{scalar} and \texttt{textual} evaluations.

Scalar metrics are primarily considered objective evaluation tools. Among them, \textit{accuracy} is one of the most fundamental metrics. We define \texttt{Critique Accuracy} to measure a model’s ability to correctly judge the validity of a given response, and \texttt{Preference Accuracy} to evaluate how well the model selects the better response from a pairwise comparison.

Textual critique, while inherently difficult to assess objectively due to its open-ended nature, remains critically important. A common approach involves conducting subjective analyses on a set of representative cases. However, such case studies are impractical for evaluating large-scale datasets. To address this limitation, we propose transforming subjective evaluation into an approximate objective assessment. Specifically, we employ GPT-4o to generate reliable reference critiques, anchored at a score of 8, which serve as pivots to guide the judge model in evaluating textual critiques. In this way, textual critiques can also be scored, denoted as \texttt{Critique Score}, and the reliability of these scores is empirically validated in our experiments.

\subsection{Objective and Subjective Evaluation}
\textbf{Critique Accuracy.}~The direct critique ability is to judge whether the response is correct. Thus, we define \texttt{Critique Accuracy} as the average accuracy across all samples, formulated as:
\begin{align}
    \mathrm{ACC_{critic}} = \frac{1}{N} \sum\nolimits_{i=1}^{N} I(\hat{y}_i = y_i),
\end{align}
where $N$ is the number of samples, $\hat{y}_i$ denotes the model's judgment of correctness for the $i$-th response, $y_i$ is the ground-truth correctness label, and $I(\cdot)$ is the indicator function that returns $1$ if and only if the condition holds, and $0$ otherwise.

\textbf{Preference Accuracy.}  
We construct a subset of pairwise response samples from \modelname{} to evaluate the model's comparative ability to identify the better response between two options of differing quality.  
\texttt{Preference Accuracy} is defined as the average accuracy of correct selections across all pairwise samples, formulated as:
\begin{align}
    \mathrm{ACC_{prefer}} = \frac{1}{N} \sum\nolimits_{i=1}^{N} I(\hat{c}_i = c_i),
\end{align}
where $N$ is the number of samples, $\hat{c}_i$ denotes the model's preferred choice for the $i$-th response pair, $c_i$ points to the higher quality response in a pairwise sample, and $I(\cdot)$ is the indicator function too.

\textbf{Critique Score.}  
As mentioned above, textual critiques can be approximately and objectively assessed using a scalar metric, termed \texttt{Critique Score}, defined as $\mathrm{Score} =$:
\begin{align}
     \frac{1}{N} \sum_{i=1}^{N} \mathrm{Score}_{i}(\mathrm{critique}_{\text{LMM}}, \mathrm{critique}_{\text{reference}}),
\end{align}
where $N$ is the number of samples, $\mathrm{Score}_{i}(\cdot)$ denotes the judged score for the $i$-th critique, bounded within $[0, 10]$.  
Here, $\mathrm{critique}_{\text{reference}}$ is a high-quality reference critique anchored at a score of $8$, and $\mathrm{critique}_{\text{LMM}}$ is the model-generated critique being evaluated against the reference. With the assistance of reference critique, it is feasible to score any textual contents, namely, basic and correction textual critiques here. The judge prompt and critique prompts can be found in Figure~\ref{fig:judge-prompt}, \ref{ap:prompt_basic}, \ref{ap:prompt_correlation}, and \ref{ap:prompt_comparative}
 (Appendix~\S\ref{Apx:B}).

%% file: tables/Table_Main_Results.tex
\begin{table*}[h]
\centering
\resizebox{\textwidth}{!}{

\begin{tabular}{@{}lccccccccccc}
\toprule
\multirow{2}{*}{\textbf{Model}} 
& \multicolumn{2}{c}{\textbf{Core}} 
& \multicolumn{2}{c}{\makecell{\textbf{Core} \\ \textbf{Single-image}}} 
& \multicolumn{2}{c}{\textbf{Open}} 
& \multicolumn{2}{c}{\makecell{\textbf{Open} \\ \textbf{Single-image}}}
& \multicolumn{2}{c}{\textbf{Avg.}} \\
& $\mathrm{ACC_{critic}}$ & Score & $\mathrm{ACC_{critic}}$ & Score & $\mathrm{ACC_{critic}}$ & Score & $\mathrm{ACC_{critic}}$ & Score & $\mathrm{ACC_{critic}}$ & Score \\
\midrule

\rowcolor{pink!50}
\multicolumn{11}{c}{\textit{Proprietary Models}}\\
\textbf{o4-mini} & \multicolumn{1}{>{\columncolor{gray!20}}c}{0.896} & 7.924
& \multicolumn{1}{>{\columncolor{gray!30}}c}{0.897} & 7.952
& \multicolumn{1}{>{\columncolor{gray!30}}c}{0.906} & 7.877
& \multicolumn{1}{c}{0.856} & 7.976
& \multicolumn{1}{>{\columncolor{gray!30}}c}{\textbf{0.900}} & 7.933\\
\textbf{GPT-4o} & \multicolumn{1}{c}{0.832} & 7.499
& \multicolumn{1}{c}{0.834} & 7.429
& \multicolumn{1}{c}{0.826} & 7.807
& \multicolumn{1}{c}{0.789} & 7.637
& \multicolumn{1}{c}{0.830} & 7.503\\
\textbf{GPT-4o-mini} & \multicolumn{1}{c}{0.833} & 6.634
& \multicolumn{1}{c}{0.836} & 6.534
& \multicolumn{1}{c}{0.762} & 6.549
& \multicolumn{1}{c}{0.690} & 6.416
& \multicolumn{1}{c}{0.821} & 6.580\\
\textbf{Claude-3.7-sonnet} & \multicolumn{1}{c}{0.834} & 8.113
& \multicolumn{1}{c}{0.828} & 8.080
& \multicolumn{1}{c}{0.799} & 8.102
& \multicolumn{1}{c}{0.808} & 8.097
& \multicolumn{1}{c}{0.831} & 8.099\\
\textbf{Gemini-2.5-flash} & \multicolumn{1}{c}{0.826} & 6.495
& \multicolumn{1}{c}{0.828} & 6.460
& \multicolumn{1}{c}{0.774} & 6.500
& \multicolumn{1}{c}{0.756} & 6.340
& \multicolumn{1}{c}{0.818} & 6.474\\
\textbf{Gemini-2.5-pro} & \multicolumn{1}{c}{0.865} & \cellcolor{gray!20}8.558
& \multicolumn{1}{c}{0.865} & \cellcolor{gray!20}8.549
& \multicolumn{1}{c}{0.865} & 8.246
& \multicolumn{1}{>{\columncolor{gray!30}}c}{0.866} & 8.325
& \multicolumn{1}{c}{0.865} & \cellcolor{gray!30}\textbf{8.514}\\
\textbf{Grok-2-vision}& \multicolumn{1}{c}{0.803} & 7.523
& \multicolumn{1}{c}{0.806} & 7.490
& \multicolumn{1}{c}{0.818} & 8.066
& \multicolumn{1}{c}{0.806} & 8.274
& \multicolumn{1}{c}{0.806} & 7.600\\
\midrule
\rowcolor{green!30}
\multicolumn{11}{c}{\textit{Open-weight Models (Larger than 30B)}}\\
\textbf{Qwen2.5-vl-32b-instruct} & \multicolumn{1}{c}{0.839} & 8.208
& \multicolumn{1}{c}{0.811} & 8.138
& \multicolumn{1}{c}{0.852} & \cellcolor{gray!20}8.566
& \multicolumn{1}{c}{0.794} & \cellcolor{gray!20}8.495
& \multicolumn{1}{c}{0.829} & 8.216\\
\textbf{Qwen2.5-vl-72b-instruct} & \multicolumn{1}{c}{0.839} & 6.931
& \multicolumn{1}{c}{0.838} & 6.817
& \multicolumn{1}{c}{0.803} & 7.089
& \multicolumn{1}{c}{0.808} & 7.133
& \multicolumn{1}{c}{0.834} & 6.911\\
\textbf{Pixtral-large}& \multicolumn{1}{c}{0.828} & 7.489
& \multicolumn{1}{c}{0.836} & 7.531
& \multicolumn{1}{c}{0.804} & 7.743
& \multicolumn{1}{c}{0.845} & 7.784
& \multicolumn{1}{c}{0.830} & 7.538\\
\textbf{Llama-4-maverick}& \multicolumn{1}{c}{0.748} & 5.811
& \multicolumn{1}{c}{0.812} & 5.971
& \multicolumn{1}{c}{0.742} & 6.342
& \multicolumn{1}{c}{0.705} & 6.250
& \multicolumn{1}{c}{0.768} & 5.938\\
\midrule
\rowcolor{green!30}
\multicolumn{11}{c}{\textit{Open-weight Models (Less than 30B)}}\\
\textbf{Gemma-3-4b} & \multicolumn{1}{c}{0.508} & 5.130
& \multicolumn{1}{c}{0.590} & 5.509
& \multicolumn{1}{c}{0.546} & 6.155
& \multicolumn{1}{c}{0.611} & 6.643
& \multicolumn{1}{c}{0.546} & 5.400\\

\textbf{Qwen2.5-vl-7b} & \multicolumn{1}{c}{0.783} & 4.617
& \multicolumn{1}{c}{0.780} & 5.007
& \multicolumn{1}{c}{0.711} & 4.573
& \multicolumn{1}{c}{0.788} & 4.806
& \multicolumn{1}{c}{0.777} & 4.765\\
\textbf{Llama-3.2-11b-vision} & \multicolumn{1}{c}{0.721} & 5.185
& \multicolumn{1}{c}{0.750} & 5.093
& \multicolumn{1}{c}{0.728} & 5.179
& \multicolumn{1}{c}{0.759} & 5.351
& \multicolumn{1}{c}{0.734} & 5.161\\
\textbf{Pixtral-12b}& \multicolumn{1}{c}{0.703} & 5.201
& \multicolumn{1}{c}{0.687} & 5.289
& \multicolumn{1}{c}{0.721} & 5.700
& \multicolumn{1}{c}{0.688} & 5.759
& \multicolumn{1}{c}{0.695} & 5.302\\
\textbf{Gemma-3-12b} & \multicolumn{1}{c}{0.759} & 6.566
& \multicolumn{1}{c}{0.739} & 6.419
& \multicolumn{1}{c}{0.645} & 6.744
& \multicolumn{1}{c}{0.671} & 6.944
& \multicolumn{1}{c}{0.742} & 6.531\\
\textbf{Gemma-3-27b} & \multicolumn{1}{c}{0.804} & 7.107
& \multicolumn{1}{c}{0.773} & 6.921
& \multicolumn{1}{c}{0.720} & 7.297
& \multicolumn{1}{c}{0.744} & 7.700
& \multicolumn{1}{c}{0.783} & 7.082\\
\textbf{Llama-4-scout} & \multicolumn{1}{c}{0.757} & 5.771
& \multicolumn{1}{c}{0.767} & 5.822
& \multicolumn{1}{c}{0.797} & 6.336
& \multicolumn{1}{c}{0.818} & 6.521
& \multicolumn{1}{c}{0.768} & 5.879\\

\bottomrule
\end{tabular}
}
\caption{Main results about $\mathrm{ACC_{critic}}$ and Score on different sub-datasets.}
\label{tab:main}
\end{table*}


%% file: sections/5.Experiments.tex
\section{Evaluation and Analysis}
\label{set:experiment}
In this section, we comprehensively analyze the critique capability of representative LMMs, and the main results are in Table~\ref{tab:main}. Subsequently, we conduct a series of in-depth experiments and analyses, where several intriguing insights are revealed.

\begin{figure*}[t]
    \centering
    \includegraphics[width=0.75\linewidth]{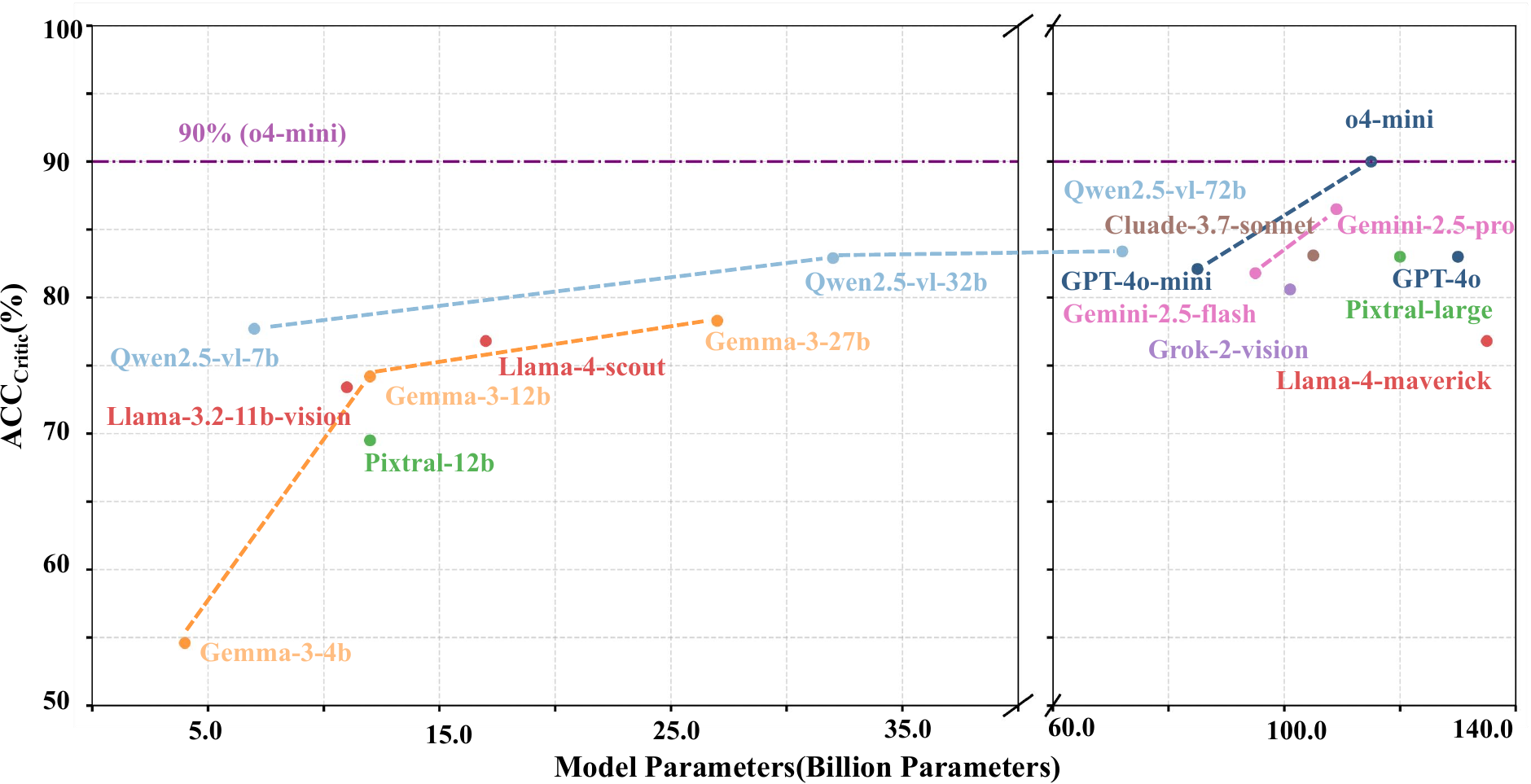}
    \caption{Scaling law on $\mathrm{ACC_{critic}}$ across models. Note that the parameter sizes of all closed-source LMMs are estimated, as their exact values are not publicly available. However, the relative scale among them is preserved — for example, Gemini-2.5-flash is known to be smaller than Gemini-2.5-pro.} \label{fig:further_scaling_acc}
\end{figure*}

\subsection{Main Results} Table~\ref{tab:main} presents the complete results across \modelname{}. Several general observations can be drawn. First, closed-source LMMs generally outperform open-source counterparts in critique performance. Notably, the o4-mini model achieves SOTA performance in terms of  $\mathrm{ACC_{critic}}$, while Gemini-2.5-pro attains the highest \texttt{Critique Score}. Second, model size (i.e., parameter scale) has a significant impact on performance.

Further exploring the experimental findings, we observe that the performance differences between the core and core single-image datasets, as well as between the open and open single-image datasets, are marginal. This may be attributed to the overlap of tasks within each sub-dataset category, leading to similar model behavior across them. In addition, the overall critique performance appears to be suboptimal when the model size is below 30 billion parameters. Among these smaller models, only Gemma-3-27B demonstrates relatively strong performance, achieving an $\mathrm{ACC_{critic}}$ of 0.783 and a critique score of 7.082. These results suggest that \textit{a model size of approximately 30 billion parameters may represent a threshold for effectively supporting LMMs’ critique capabilities.}

\textbf{Scaling Law.} 
To verify whether the scaling law holds in the context of critique evaluation in \modelname{}, Figure~\ref{fig:further_scaling_acc} visualizes the $\mathrm{ACC_{critic}}$ results across LLMs with increasing model sizes. The results clearly indicate that $\mathrm{ACC_{critic}}$ scores for models within the same series (e.g., the Gemma-3 series) consistently improve as the parameter size increases. Even among closed-source LMMs, larger models consistently outperform their smaller counterparts within the same series, e.g., Gemini-2.5-pro outperforms Gemini-2.5-flash. This indirectly supports the reliability of our critique evaluation and demonstrates the robustness of \modelname{}.

\subsection{Further Analysis}

\textbf{Effects of Task Type.}~While Table~\ref{tab:main} presents the overall results for each sub-dataset, \modelname{}, as a comprehensive benchmark, covers a diverse range of tasks categorized into eight primary task types. Therefore, it is crucial to examine model performance across these distinct task categories to gain deeper insights. Appendix~\S\ref{Apx:c} provides detailed model performance results across the eight task types for each sub-dataset in Table~\ref{tab:core_socre},\ref{tab:core-single-score},\ref{tab:open-socre},\ref{tab:open-single-score},\ref{tab:core-acc},\ref{tab:core-single-acc},\ref{tab:open-acc}, and \ref{tab:open-single-acc}. These detailed results indicate that the overall SOTA models also maintain strong performance across all task types. Notably, Claude-3.7-Sonnet, as a high-performing LMM, consistently achieves top results on coding tasks across each sub-dataset.

\textbf{Multiple Critique Dimensions.}~To more effectively evaluate a model's self-improvement capability, we introduce two additional critique dimensions: \textit{correctness critique} and \textit{comparative critique}. The former assesses the model’s ability to identify and correct errors in corresponding responses, while the latter evaluates the model’s capacity to select the better response from a pair of differing-quality answers.

\input{tables/Table_Correctness}
\input{tables/Table_Compare_ACC}

Based on the main results in Table~\ref{tab:main}, we select representative and high-performing open- and closed-source LMMs for further in-depth experiments. Table \ref{tab:correlation} shows that the closed-source model Gemini-2.5-pro achieves the highest score in the correctness critique, which is consistent with its overall performance in the main results. Notably, the average correctness critique scores across models are generally lower than their corresponding critique scores in the main evaluation, suggesting that \textit{correctness critique poses a greater challenge}.

Table~\ref{tab:comp} reveals two key findings: First, among the closed- and open-source LMMs, Gemini-2.5-pro and Llama-4-maverick demonstrate the strongest performance. Second, the (medium, high) pairwise sub-dataset is evidently the most difficult, likely \textit{due to the subtle differences in quality between medium and high responses, making preference judgment more challenging}.

\textbf{Effects of Response Quality.}~In \modelname{}, we employed GPT-4o to score the quality of all generated responses. This naturally motivates an exploration of the relationship between response quality and the corresponding critique scores.

\begin{figure*}[t]
    \centering
    \includegraphics[width=0.75\linewidth]{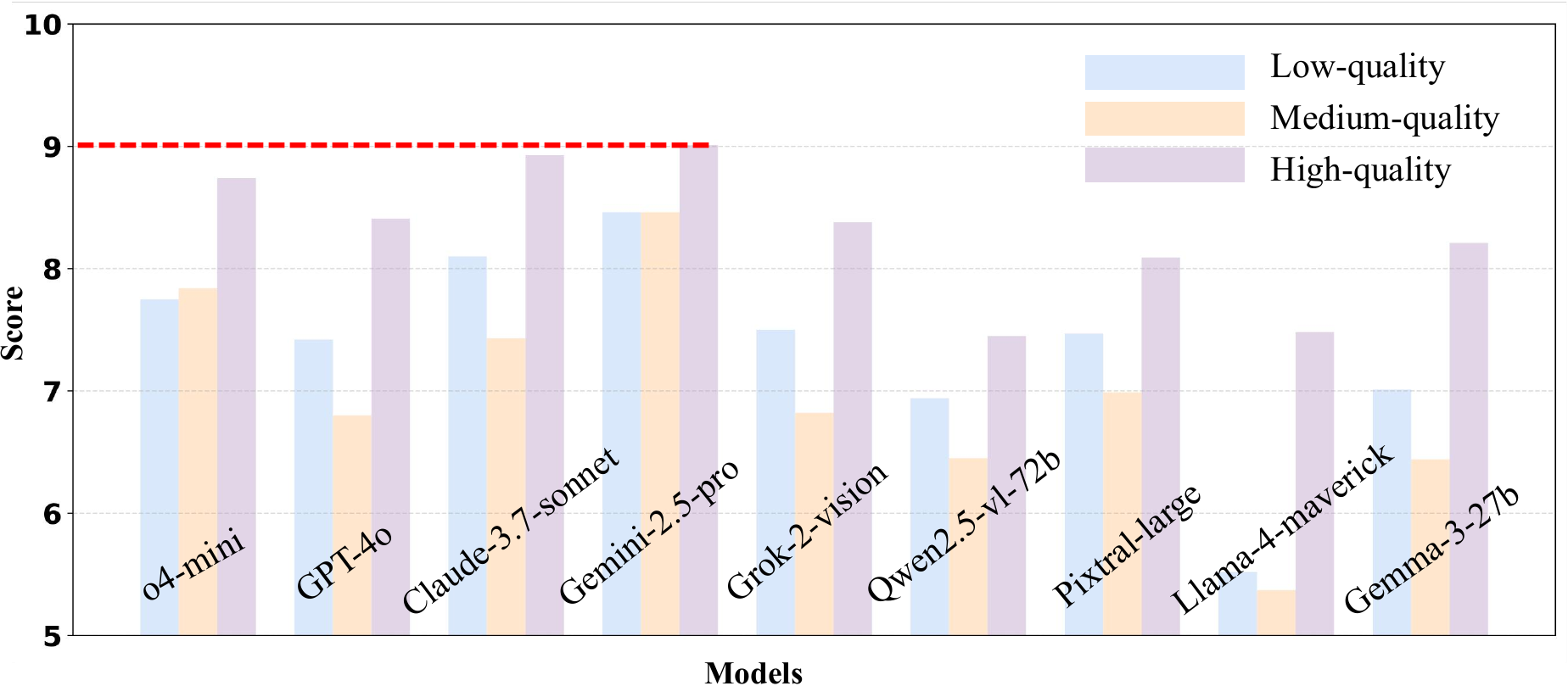}
    \caption{The distribution of critique scores across responses of different quality levels, where low-, medium-, and high-quality correspond to labeled response quality ranges of $[0,4]$, $[5,7]$, and $[8,10]$, respectively.
} \label{fig:quality-score}
\end{figure*}

Figure \ref{fig:quality-score} reveals some intriguing insights: high-quality responses tend to get high critique scores. Another interesting observation is that medium-quality responses are the most challenging, producing the lowest critique scores among the three groups. This result aligns with intuitive reasoning, as responses that are either good or poor exhibit more distinct characteristics, whereas medium-quality responses pose greater challenges for critique. \textit{This provides a direction for further enhancing model performance, specifically by focusing more on medium-quality responses.}

\textbf{Reliability and Bias of Subjective Evaluation.}~In utilizing a judge model for subjective evaluation, it is crucial to ensure the reliability of its assessments and to reduce the discrepancy between human and model judgments. To this end, we deliberately designed scoring rubrics grounded in expert human reasoning (Table~\ref{tab:rubric}), and employed GPT-4o to generate reference critiques based on these rubrics, with access to the ground-truth answers. When utilizing a judge model to evaluate LMMs' critique scores, the evaluation reliability can be significantly improved by providing a corresponding reference critique anchored at a score of 8. 

\begin{figure}[t]
    \centering
    \includegraphics[width=\linewidth]{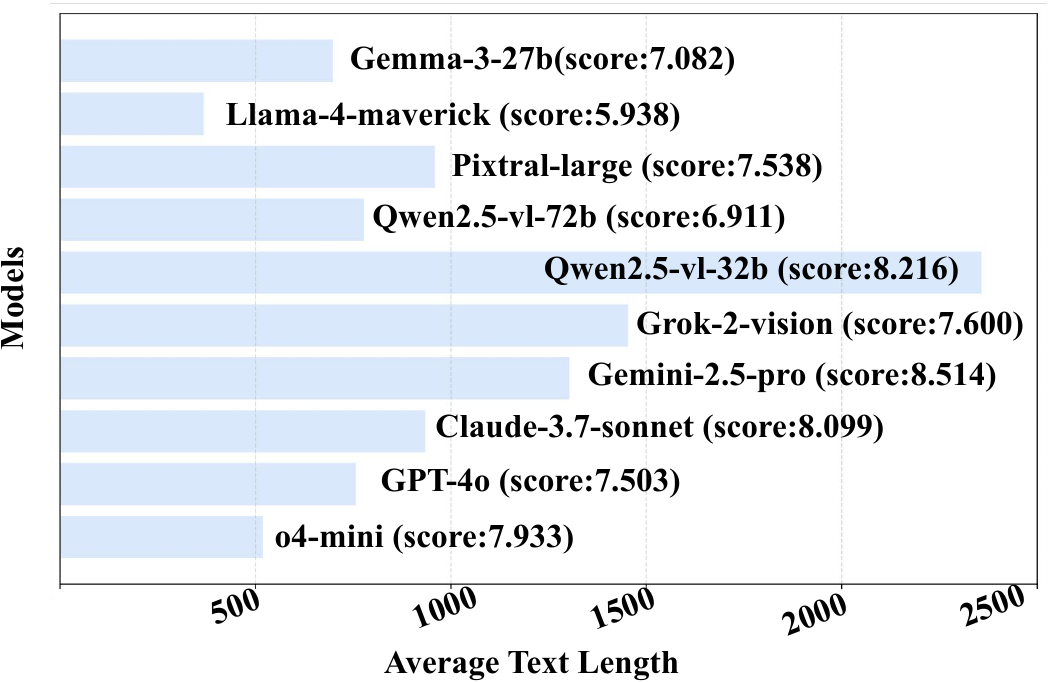}
    \caption{The relationship between the average length of textual critiques and critique scores across models.} \label{fig:text-score}
\end{figure}

\textit{Why do we need reference critiques?} In Figure~\ref{fig:case1} (Appendix~\S\ref{Apx:D}), we present a mathematical reasoning task in which the model-generated response is incorrect. The critique model, o4-mini, successfully identifies this error and provides a comprehensive textual critique, including detailed reasoning steps and a counterexample. When evaluated by the judge model, the critique is recognized as superior to the reference critique—particularly due to the inclusion of the counterexample—and is assigned a higher score of 9, compared to the reference critique's anchored score of 8. This case study demonstrates that the use of a reference critique effectively guides the judge model’s assessment, thereby enhancing the reliability of the evaluation compared to scoring without such a reference.

\textit{Why does the Judge model always exhibit evaluation bias?}
As mentioned above, the use of reference critiques can effectively enhance the reliability of model-based evaluations. However, discrepancies between model and human judgments inevitably persist. Therefore, it is crucial to conduct an in-depth analysis of the potential factors contributing to this bias.

After examining a large number of cases, we observed an emerging pattern: \textit{the critique score appears to be positively correlated with the length of the textual critique}. As shown in Figure~\ref{fig:text-score}, none of the models achieve a critique score exceeding 8 when their average text length is below 1000.

Besides, an unexpected observation emerges between Qwen2.5-vl-32b and Qwen2.5-vl-72b: the smaller model outperforms the larger one in terms of critique score. However, this result becomes more interpretable when considered from the perspective of textual length. Since longer critiques often entail more comprehensive, step-by-step reasoning, we find that the judge model tends to assign higher scores to such responses.

\textbf{Potential Bias Caused by Model Style.} To assess potential bias from a specific model style~(e.g., GPT-based series), we constructed a sub-dataset from MM-CRITIC. Reference critiques were generated by Gemini-2.5-flash, and scoring was judged by GPT-4.1 and Claude-4.0-sonnet, respectively. As the Table~\ref{tab:ablation} in Appendix~\ref{Apx:ablation} shown, five models were evaluated: o4-mini, GPT-4o, GPT-4o-mini, Claude-3.7-sonnet, and Gemini-2.5-pro. Results show that model rankings remain consistent with those judged by GPT-4.1 with reference critiques annotated by GPT-4o. Only o4-mini and Claude-3.7-sonnet swapped ranks when judging by Claude-4.0-sonnet, which is acceptable given their small score gap (0.166) in the main experiments (in Table~\ref{tab:ablation}). These findings suggest that \textit{GPT-4o does not significantly affect the fairness or validity of our evaluation}.

\textbf{Case study.}~Case studies, particularly those involving poor-performing examples, can provide valuable insights into the limitations of both the evaluation methodology and the critique capabilities of the models. We present representative cases to facilitate in-depth analysis in Appendix~\S\ref{Apx:D}.

\textit{Instruction following or formulaic step-by-step reasoning?}~A key finding is the conflict between following instructions and generating detailed reasoning. While prompts request brief, direct answers, some LMMs consistently produce step-by-step explanations, ignoring instructions. Though such reasoning can improve response quality, it may not match user expectations for concise replies, the case can be found in Figure~\ref{fig:case2} (Appendix~\S\ref{Apx:D}).

%% file: tables/Table_Correctness.tex
\begin{table*}[h]
\centering
\resizebox{\textwidth}{!}{

\begin{tabular}{@{}lccccccccccc}
\toprule
\multirow{2}{*}{\textbf{Model}} 
& \multicolumn{9}{c}{\textbf{Task types}} 
\\
\cline{2-9} 
& \textbf{Perception } & \textbf{Planning} & \textbf{Knowledge } &\textbf{Information Extraction} & \textbf{Mathematics} & \textbf{Coding} & \textbf{Science} & \textbf{Metric} & \textbf{Avg.} \\
\midrule

\rowcolor{pink!50}
\multicolumn{10}{c}{\textit{Proprietary Models}}\\
\textbf{o4-mini} & \multicolumn{1}{c}{5.636} & 6.097
& \multicolumn{1}{c}{6.290} & 7.625
& \multicolumn{1}{c}{5.171} & 7.000
& \multicolumn{1}{c}{6.457} & 6.324
& \multicolumn{1}{c}{6.220} \\
\textbf{GPT-4o} & \multicolumn{1}{c}{5.606} & 6.129
& \multicolumn{1}{c}{5.323} & 5.875
& \multicolumn{1}{c}{5.114} & 6.424
& \multicolumn{1}{c}{6.500} & 6.794
& \multicolumn{1}{c}{5.980} \\
\textbf{Claude-3.7-sonnet} & \multicolumn{1}{>{\columncolor{gray!20}}c}{7.406} & 7.267
& \multicolumn{1}{c}{6.839} & 8.375
& \multicolumn{1}{c}{6.114} & 5.969
& \multicolumn{1}{c}{7.943} & 7.100
& \multicolumn{1}{c}{7.041} \\
\textbf{Gemini-2.5-pro} & \multicolumn{1}{c}{7.152} & \cellcolor{gray!20}7.630
& \multicolumn{1}{>{\columncolor{gray!20}}c}{7.152} & \cellcolor{gray!20}8.875
& \multicolumn{1}{>{\columncolor{gray!20}}c}{7.852} & \cellcolor{gray!20}7.471
& \multicolumn{1}{>{\columncolor{gray!20}}c}{8.667} & \cellcolor{gray!20}7.735
& \multicolumn{1}{>{\columncolor{gray!30}}c}{\textbf{7.794}} \\
\textbf{Grok-2-vision}& \multicolumn{1}{c}{7.152} & 5.103
& \multicolumn{1}{c}{5.516} & 3.188
& \multicolumn{1}{c}{4.857} & 3.737
& \multicolumn{1}{c}{4.600} & 5.242
& \multicolumn{1}{c}{5.113} \\
\midrule
\rowcolor{green!30}
\multicolumn{10}{c}{\textit{Open-weight Models}}\\

\textbf{Qwen2.5-vl-72b} & \multicolumn{1}{c}{6.458} & 4.500
& \multicolumn{1}{c}{5.074} & 3.000
& \multicolumn{1}{c}{4.448} & 5.294
& \multicolumn{1}{c}{6.514} & 6.500
& \multicolumn{1}{c}{5.486} \\
\textbf{Pixtral-large} & \multicolumn{1}{c}{6.625} & 4.690
& \multicolumn{1}{c}{5.111} & 6.200
& \multicolumn{1}{c}{4.100} & 4.909
& \multicolumn{1}{c}{6.118} & 6.333
& \multicolumn{1}{c}{5.410}\\
\textbf{Llama-4-maverick}& \multicolumn{1}{c}{3.225} & 3.452
& \multicolumn{1}{c}{3.100} & 5.500
& \multicolumn{1}{c}{1.471} & 4.600
& \multicolumn{1}{c}{2.324} & 1.941
& \multicolumn{1}{c}{2.970}\\
\textbf{Gemma-3-27b} & \multicolumn{1}{c}{7.061} & 6.065
& \multicolumn{1}{c}{6.258} & 4.813
& \multicolumn{1}{c}{5.400} & 4.381
& \multicolumn{1}{c}{7.114} & 7.242
& \multicolumn{1}{>{\columncolor{gray!30}}c}{6.217} \\

\bottomrule
\end{tabular}
}
\caption{Correction critique scores on different task types.}
\label{tab:correlation}
\end{table*}

%% file: tables/Table_Compare_ACC.tex
\begin{table}[h]
\small
\centering

\begin{tabular}{@{}lccc@{}}
\toprule
\multirow{2}{*}{\textbf{Model}} & \multicolumn{3}{c}{$\mathrm{ACC_{prefer}}$} \\
\cline{2-4}
& \textbf{G1} & \textbf{G2} & \textbf{G3} \\
\hline

\rowcolor{pink!50}
\multicolumn{4}{c}{\textit{Proprietary Models}} \\
\textbf{o4-mini} & 0.836 & 0.658 & 0.831 \\
\textbf{GPT-4o} & 0.848 &  0.589 & 0.740 \\
\textbf{Claude-3.7-sonnet} & 0.835 & 0.579 & 0.785 \\
\textbf{Gemini-2.5-pro} & 0.860 & \cellcolor{gray!30}0.716 &  \cellcolor{gray!30}0.939 \\
\textbf{Grok-2-vision} &  \cellcolor{gray!30}0.867 & 0.475 &  0.687 \\
\midrule
\rowcolor{green!30}
\multicolumn{4}{c}{\textit{Open-weight Models}} \\
\textbf{Qwen2.5-vl-72b} & 0.733 & 0.507 & 0.696 \\
\textbf{Pixtral-large} & 0.858 & 0.542 & 0.744 \\
\textbf{Llama-4-maverick} & 0.854 & \cellcolor{gray!30}0.658 & \cellcolor{gray!30}0.821 \\
\textbf{Gemma-3-27b} &  \cellcolor{gray!30}0.856 & 0.615 & 0.757 \\

\bottomrule
\end{tabular}
\caption{$\mathrm{ACC_{prefer}}$ results across models on the three comparative pairwise sub-datasets, where \textbf{G1}, \textbf{G2}, and \textbf{G3} represent the response quality combinations of (low, medium), (medium, high), and (low, high), respectively.}
\label{tab:comp}
\end{table}

%% file: sections/6.Conclusion.tex
\section{Conclusion}
\label{set:conc}
In this paper, we introduce \modelname{}, a holistic and reliable benchmark for evaluating the critique abilities of LMMs across multiple dimensions. Extensive experiments demonstrate the basic critique performance of leading LMMs and validate the reliability of \modelname{} through the observed scaling law. Further analysis reveals valuable insights, including the correlation between response quality and critique scores, varying levels of critique difficulty across dimensions, and potential judgment biases linked to critique text richness. Our \modelname{} offers a solid foundation for benchmarking and advancing the critique capabilities of LMMs, fostering the development of more explainable and trustworthy multimodal systems.

%% file: sections/7.appendix.tex
\appendix
\onecolumn
\section{Dataset and LMMs Information.}
\label{Apx:A}

\begin{table*}[h]\label{tab:data-collection}
\centering
\resizebox{\textwidth}{!}{
\begin{tabular}{@{}cccccc@{}}
\toprule
 \multirow{2}{*}{\textbf{Task Type}}& \multicolumn{4}{c}{\textbf{Sub-datasets}} &\\ \cline{2-6}
  & \textbf{Core} &\textbf{Core Single-image}& \textbf{Open} & \textbf{Open Single-image}& \textbf{Total} \\ 
  \midrule
 \textbf{Perception}&266 &146 & 24&18&454 \\
 \textbf{Planning}& 146& 80& 10& 8&244\\
 \textbf{Knowledge}&142 &108 & 52&46&348 \\
 \textbf{Information Extraction}& 112&78 &32 & 4&226\\
 \textbf{Mathematics}&66 &60 &- &-&126 \\
 \textbf{Coding} &58 &28 &4 &4&94 \\
 \textbf{Science}&54& 40&4 & 4&102\\
 \textbf{Metric} &36 &6 & 4&-&46 \\
 \midrule
 \textbf{Total}&880 & 546&130 & 84&1640\\
\bottomrule
\end{tabular}
}
\caption{The statistics of tasks that are selected to generate responses for our benchmark.}
\label{tab:data-colllect}
\end{table*}

\begin{table*}[h]
\centering
\resizebox{\textwidth}{!}{
\begin{tabular}{@{}cccccc@{}}
\toprule
 \multirow{2}{*}{\textbf{Task Type}}& \multicolumn{4}{c}{\textbf{Sub-datasets}} &\\ \cline{2-6}
  & \textbf{Core} &\textbf{Core Single-image}& \textbf{Open} & \textbf{Open Single-image}& \textbf{Total} \\ 
  \midrule
 \textbf{Perception}&668 &435 & 61& 54& 1218 \\
 \textbf{Planning}& 320& 238& 26& 23& 607\\
 \textbf{Knowledge}&394 &319 & 150&137& 1000 \\
 \textbf{Information Extraction}& 290&231 &51 & 12&584\\
 \textbf{Mathematics}&189 &179 &- &-&368 \\
 \textbf{Coding} &170 & 82 &12 &12&276 \\
 \textbf{Science}&161& 118 &12 & 12&303\\
 \textbf{Metric} &90 & 18 & 7 &-&115 \\
 \midrule
 \textbf{Total}&2282 & 1620&319 & 250&4471\\
\bottomrule
\end{tabular}
}
\caption{Dataset statistics of \modelname{}.}
\label{tab:data-response-collect}
\end{table*}

\begin{table*}[h]
\centering
\resizebox{0.8\textwidth}{!}{
\begin{tabular}{@{}ccccc@{}}
\toprule
 \multirow{2}{*}{\textbf{Task Type}}&\multirow{2}{*}{\textbf{Correction }}& \multicolumn{3}{c}{\textbf{Comparison}}\\ 
 \cline{3-5}
  &  &\textbf{Group 1}& \textbf{Group 2} & \textbf{Group 3} \\ 
  \midrule
 \textbf{Perception} &35&30 &30 &30\\
 \textbf{Planning}&35&30 &1 &23\\
 \textbf{Knowledge}& 35&30 &16 &30\\
 \textbf{Information Extraction}&35&30 &13& 30\\
 \textbf{Mathematics} &35&22 & -& 24\\
 \textbf{Coding} &35&30 &7&15 \\
 \textbf{Science}&35&25 &5&26\\
 \textbf{Metric} &35&11 &4&23\\
 \midrule
 \textbf{Total} &280 &208&76&201\\
\bottomrule
\end{tabular}
}
\caption{Dataset statistics of sub-datasets for correction and comparative critique in \modelname{}.}
\label{tab:data-subset-corec-compa}
\end{table*}

\begin{table*}[ht]
\small
    \centering
    
    \resizebox{\textwidth}{!}{
        \begin{tabular}{@{}ll@{}}
        \toprule
        \textbf{LMMs}   & \textbf{Source} \\ \midrule
         \textbf{InternVL2.5-4B} & \url{https://huggingface.co/OpenGVLab/InternVL2_5-4B}\\ 
        \textbf{InternVL2.5-8B } & \url{https://huggingface.co/OpenGVLab/InternVL2_5-8B}\\ 
        \textbf{InternVL2.5-26B } & \url{https://huggingface.co/OpenGVLab/InternVL2_5-26B}   \\
         \textbf{Phi-3.5-vision-instruct} & \url{https://huggingface.co/microsoft/Phi-3.5-vision-instruct}\\ 
        \textbf{Phi-4-multimodal-instruct} & \url{https://huggingface.co/microsoft/Phi-4-multimodal-instruct} \\ 
        \textbf{Qwen2.5-vl-3b-Instruct} & \url{https://huggingface.co/Qwen/Qwen2.5-vl-3b-Instruct}\\ 
        \textbf{Qwen2.5-vl-7b-Instruct} & \url{https://huggingface.co/Qwen/Qwen2.5-vl-7b-Instruct}\\ 
        \textbf{Deepseek-vl2-tiny} & \url{https://huggingface.co/deepseek-ai/deepseek-vl2-tiny} \\ 
        \textbf{Llava-1.5-7b-hf} & \url{https://huggingface.co/llava-hf/llava-1.5-7b-hf} \\ 
        \textbf{Llava-onevision-qwen2-7b-ov-hf} & \url{https://huggingface.co/llava-hf/llava-onevision-qwen2-7b-ov-hf} \\ 
        \textbf{Llama-3.2-11b-vision-Instruct} & \url{https://huggingface.co/unsloth/Llama-3.2-11b-vision-Instruct} \\ 
        \textbf{Pixtral-12b} & \url{https://huggingface.co/mistral-community/pixtral-12b} \\ 
         \hline
        
        \end{tabular}
    }
    \caption{The list of used LMMs for generating responses.}
    \label{tab:llms_list}
\end{table*}

\begin{table*}[ht]
\small
    \centering    
    \resizebox{\textwidth}{!}{
        \begin{tabular}{@{}ll@{}}
        \toprule
        \textbf{LMMs}   & \textbf{Source} \\ \midrule
         \textbf{openai/o4-mini} & \url{https://openrouter.ai/openai/o4-mini}\\ 
        \textbf{openai/gpt-4o-2024-11-20 } & \url{https://openrouter.ai/openai/gpt-4o-2024-11-20www}\\ 
        \textbf{openai/gpt-4o-mini} & \url{https://openrouter.ai/openai/gpt-4o-mini}   \\
         \textbf{anthropic/claude-3.7-sonnet} & \url{https://openrouter.ai/anthropic/claude-3.7-sonnet}\\ 
        \textbf{google/gemini-2.5-flash-preview} & \url{https://openrouter.ai/google/gemini-2.5-flash-preview} \\ 
        \textbf{google/gemini-2.5-pro-preview} & \url{https://openrouter.ai/google/gemini-2.5-pro-preview}\\ 
        \textbf{x-ai/grok-2-vision-1212} & \url{https://openrouter.ai/x-ai/grok-2-vision-1212}\\ 
        \textbf{qwen/qwen2.5-vl-32b-instruct} & \url{https://openrouter.ai/qwen/qwen2.5-vl-32b-instruct} \\ 
        \textbf{qwen/qwen2.5-vl-72b-instruct} & \url{https://openrouter.ai/qwen/qwen2.5-vl-72b-instruct} \\ 
        \textbf{mistralai/pixtral-large-2411} & \url{https://openrouter.ai/mistralai/pixtral-large-2411} \\ 
        \textbf{meta-llama/llama-4-maverick} & \url{https://openrouter.ai/meta-llama/llama-4-maverick} \\ 
        \textbf{qwen/qwen-2.5-vl-7b-instruct} & \url{https://openrouter.ai/qwen/qwen-2.5-vl-7b-instruct} \\ 
        \textbf{meta-llama/llama-3.2-11b-vision-instruct} & \url{https://openrouter.ai/meta-llama/llama-3.2-11b-vision-instruct} \\ 
        \textbf{mistralai/pixtral-12b} & \url{https://openrouter.ai/mistralai/pixtral-12b} \\
        \textbf{google/gemma-3-12b-it} & \url{https://openrouter.ai/google/gemma-3-12b-it} \\ 
        \textbf{google/gemma-3-27b-it} & \url{https://openrouter.ai/google/gemma-3-27b-it} \\ 
        \textbf{meta-llama/llama-4-scout} & \url{https://openrouter.ai/meta-llama/llama-4-scout} \\ 
        \textbf{google/gemma-3-4b-it} & \url{https://openrouter.ai/google/gemma-3-4b-it} \\ 
        
         \hline
        
        \end{tabular}
    }
    \caption{The list of used LMM APIs through OpenRouter.}
    \label{tab:llms_inference_list}
\end{table*}

\clearpage
\section{Prompts and Scoring Rubric Checklist}
\label{Apx:B}

\begin{table*}[h]
\small
\centering
\label{tab:rubric_for_domains}
\resizebox{1.0\textwidth}{!}{
    \begin{tabular}{@{}cl@{}}
    \toprule
    \textbf{Task}   & \textbf{Score Rubric} \\ \midrule

     \textbf{Common for All Tasks} & \begin{tabular}[c]{@{}l@{}}\textbf{Correctness}: \\ For tasks with ground truth, carefully check the response whether gives correct answer; \\ For tasks with open answers, carefully analyze the accuracy of generated responses, \\ including but not limited to the following aspects:  \\ $\star$ consistent with reference answer\\  $\star$ factual knowledge\\ 
     
    \textbf{Response Quality}: Carefully analyze the quality of generated responses, \\ including but not limited to the following aspects: \\ $\star$ correct spelling/grammar \\ $\star$ readability and comprehensibility\\  $\star$ effectiveness or usefulness.\\ \end{tabular}\\\hline
    
    \textbf{Knowledge} & \begin{tabular}[c]{@{}l@{}}\textbf{Factuality}: To check the generated response whether it is in line with facts. \\ If a response is based on a false premise, it can be regarded as a bad sample.  \\
    \textbf{Relevance}: Consider whether the generated content is relevant to the question. \\ If the content is unrelated to the question, it can be reviewed as low quality. 
    \end{tabular}\\\hline
    
    \textbf{Perception} & \begin{tabular}[c]{@{}l@{}}
    \textbf{Detail}: This criterion aims to check whether the generated content contains sufficient and correct detail. \\ A response is considered lower quality if it is overly brief and lacks details.\end{tabular}\\ \hline

    \textbf{Information Extraction} & \begin{tabular}[c]{@{}l@{}}\textbf{Effectiveness}: This criterion aims to check whether the answers effectively extract information,\\ based on the question. If the generated answers do not provide effective information, \\ they can be regarded as lower quality.  \\
    \end{tabular}\\ \hline
    
    \textbf{Planning} & \begin{tabular}[c]{@{}l@{}} \textbf{Feasibility:} It is vital to assess whether the generated planning is feasible in the real world. \\ If the planning is unfeasible, it can be viewed as lower quality.
    
    \end{tabular} \\ \hline
    
    \textbf{Science} & \begin{tabular}[c]{@{}l@{}} \textbf{Factuality}: To check whether the generated response is in line with scientific facts. \\ If a response is based on a false premise, it can be regarded as a bad sample.  \\ \end{tabular}\\ \hline

    \textbf{Metric} & \begin{tabular}[c]{@{}l@{}} \textbf{Effective Utilization}: To check whether the generated response is in line with scientific facts. \\ If a response is based on a false premise, it can be regarded as a bad sample.  \\ \end{tabular}\\ \hline

    \textbf{Mathematics} & \begin{tabular}[c]{@{}l@{}} \textbf{Correctness}: Correctness-based for mathematics is a multi-step checking criterion, \\firstly assess the generated response whether it provides a correct reasoning process (if it includes),\\ Secondly check the generated response whether it provides the correct answer. \\ 
    If the generated response provides inappropriate reasoning and a wrong answer, \\ it can be regarded as of very poor quality. \\  
    If the generated response provides correct reasoning and a wrong answer, \\ it can be regarded as of relatively high quality, compared with the above case. \\  
    \end{tabular}\\ \hline

    \textbf{Coding} & \begin{tabular}[c]{@{}l@{}} \textbf{Program Grammar}: This criterion aims to check whether the generated codes \\ align with specific program language features. \\ If the generated codes utilize non-existent program language features, \\ they can be regarded as of low quality. \\
    \textbf{Correctness}: Correctness-based for coding is a multi-step checking criterion, \\firstly assess the generated response whether it provides a correct coding framework (if it includes),\\ Secondly check the generated response whether it provides correct output of codes. \\ 
    If the generated response provides an unreasonable coding framework and a wrong output, \\ it can be regarded as of very poor quality. \\  
    If the generated response provides both a correct coding framework and output, \\ it can be regarded as of relatively high quality, compared with the above case.   \\ \end{tabular}\\ \hline
    
    \hline
    \end{tabular}
}
\caption{The score rubrics for different task types. Human experts use these score rubrics to check and annotate.}
\label{tab:rubric}
\end{table*}

\begin{figure*}[h!]
\begin{center}
    \fontsize{8.4}{8.4} \selectfont
    \begin{tcolorbox}[width=1\textwidth, colback=lightblue, title={\textbf{Basic Reference Critique Generation Prompt for GPT-4o}}]

    You are a professional critical AI specialist who can evaluate the response generated by a vision large language model with corresponding domain knowledge of questions. You need to refer to the following rubrics:\\
    
    \textbf{\#Common for All Tasks}:\\
    \textbf{\#\#Correctness}: \\ For tasks with ground truth, carefully check the response whether gives correct answer; \\ For tasks with open answers, carefully analyze the accuracy of generated responses, \\ including but not limited to the following aspects:  \\ $\star$ consistent with reference answer\\  $\star$ factual knowledge.
     
    \textbf{\#\#Response Quality}: Carefully analyze the quality of generated responses, \\ including but not limited to the following aspects: \\ $\star$ correct spelling/grammar \\ $\star$ readability and comprehensibility\\  $\star$ effectiveness or usefulness. \\
    
   \textbf{\#} Besides, the question is about \texttt{application\_name}, you also need to carefully refer to the emphasized rubrics: \\
    \textcolor{c1}{(Corresponding rubric for different task types, here is an example for \texttt{Knowledge})} \\
    \textbf{\#\#Factuality}: To check whether the generated response is in line with facts. \\ If a response is based on a false premise, it can be regarded as a bad sample.  \\
    \textbf{\#\#Relevance}: Consider whether the generated content is relevant to the question. \\ If the content is unrelated to the question, it can be reviewed as low quality. \\
    \textbf{------------------------------------------------------------------------------------------------------------------------}\\
    \textbf{\#}The following is the \texttt{question} and the \texttt{response} generated by a vision large language model: \\ 
    --- Start of Question ---\\
    \textcolor{ora}{\$Question (include images)\$}\\
    --- End of Question ---\\
    
    --- Start of Response ---\\
    \textcolor{ora}{\$Response\$}\\
    --- End of Response ---

     \textbf{------------------------------------------------------------------------------------------------------------------------}\\

     \textbf{\#}Here is the \texttt{ground truth answer} (or \texttt{reference answer}), \\ which can effectively help you give reliable evaluations about the response:\\
     --- Start of Answer ---\\
    \textcolor{ora}{\$Answer\$}\\
    --- End of Answer ---
    
   \textbf{------------------------------------------------------------------------------------------------------------------------}\\

    \textbf{\#Evaluation Steps}:\\
    \textbf{\#\#}First, you need to score the response quality, and the score ranges from 0 to 10 as an integer, \\
    -[0,3] corresponds to a low-quality response,\\
    -[4,7] corresponds to a medium-quality response,\\
    -[8,9] corresponds to a high-quality response,\\
    -10 corresponds to a correct response.\\
    \textbf{\#\#} Second, you need to give a textual critique including but not limited to the following  requirements:\\
    - Provide detailed, point-by-point feedback on the answer.\\
    - Each critique should be specific and self-contained.\\
    - Clearly identify any issues, avoiding vague or ambiguous descriptions.\\
    - Offer constructive suggestions for improvement.\\
    \textbf{\#Output Format}:\\
    Provide the evaluation in JSON format as follows:\\
    \verb|```|json\\
    \{\\
        \quad"correct": "Based on the ground truth answer (if have), indicate whether the assistant's response is ['Correct', 'Error']"\\
        \quad"response\_quality": "A specific integer score ranging from 0 to 10 ",\\
        \quad"reference\_critique": "Based on the evaluation, give a comprehensive textual critique"\\
        \quad"reference\_correct": "Based on the evaluation, give a modification if the response is not of good quality enough."\\
    \}\\
    \verb|```|\\

    \end{tcolorbox}
\end{center}
\caption{Basic Reference Critique Generation Prompt for GPT-4o.} \label{ap:prompt_level1}
\end{figure*}

\begin{figure*}[h!]
\begin{center}
    \fontsize{8.4}{8.4} \selectfont
    \begin{tcolorbox}[width=1\textwidth, colback=lightblue, title={\textbf{Critique evaluation judge prompt with reference critique.}}]

    You are a professional critique evaluation judge who can evaluate the critique generated by a vision large language model based on the corresponding question and response. \\

    \textbf{\#}: The following are the \texttt{question} and generated \texttt{response}, and \texttt{critique/correct} that need to be evaluated, respectively.  \\
    \textbf{------------------------------------------------------------------------------------------------------------------------}

    --- Start of Question and Response ---\\
    \textcolor{ora}{\$Question (include images)\$}\\
    \textcolor{ora}{\$Response\$}\\
    --- End of Question and Response ---\\

     --- Start of Critique ---\\
    \textcolor{ora}{\$Critique/Correct\$}\\
    --- End of Critique ---

     \textbf{------------------------------------------------------------------------------------------------------------------------}

     \textbf{\#}Here is the \texttt{reference critique/correct}:\\
     --- Start of Reference Critique ---\\
    \textcolor{ora}{\$Reference Critique/Correct\$}\\
    --- End of Reference Critique ---
    
   \textbf{------------------------------------------------------------------------------------------------------------------------}\\

    \textbf{\#Very important rules !!!}:\\
    \textbf{\#\#}1. If you think the generated critique is better than the reference critique, you could provide a score higher than 8 \\
    \textbf{\#\#}2. Please first generate your explanation for the generated critique by comparing it with the reference critique, and then you must generate the score, where the score ranges from 0 to 10 as an integer.

    \textbf{\#Output Format}:\\
    Provide the evaluation in JSON format as follows:\\
    \verb|```|json\\
    \{\\
        \quad"explanation": "Based on the reference critique, give explanation about the score given by yourself", \\
        \quad"score": "A specific integer score ranging from 0 to 10 ",\\
    \}\\
    \verb|```|\\

    \end{tcolorbox}
\end{center}
\caption{Critique evaluation judge prompt with reference critique.} \label{ap:prompt_judge}
\label{fig:judge-prompt}
\end{figure*}


\begin{figure*}[h!]
\begin{center}
    \fontsize{8.4}{8.4} \selectfont
    \begin{tcolorbox}[width=1\textwidth, colback=lightblue, title={\textbf{Basic critique prompt.}}]

    You are a professional expert in critique who can evaluate the responses generated by large multimodal models and analyze response quality. \\

    \textbf{\#}: The following are the \texttt{question} and the generated \texttt{response} from large multimodal models.  \\
    \textbf{------------------------------------------------------------------------------------------------------------------------}

    --- Start of Question ---\\
    \textcolor{ora}{\$Question (include images)\$}\\
    
    --- End of Question---\\

     --- Start of Response  ---\\
    \textcolor{ora}{\$Response \$}\\
    --- End of Response ---\\

   \textbf{------------------------------------------------------------------------------------------------------------------------}\\

    \textbf{\#Very important rules !!!}:\\
    \textbf{\#\#}1. Carefully understand the question, and then analyze the response. \\
    \textbf{\#\#}2. Please first generate your judgment of correctness about the response, and then give a comprehensive textual critique to explain.

    \textbf{\#Output Format}:\\
    Provide the evaluation in JSON format as follows:\\
    \verb|```|json\\
    \{\\
        \quad"correct": "Based on the ground truth answer (if have), indicate whether the assistant's response is ['Correct', 'Error']",\\
        \quad"critique": "Give a comprehensive textual critique, due to the question and response", \\
        
    \}\\
    \verb|```|\\

    \end{tcolorbox}
\end{center}
\caption{Basic critique prompt.} \label{ap:prompt_basic}
\end{figure*}


\begin{figure*}[h!]
\begin{center}
    \fontsize{8.4}{8.4} \selectfont
    \begin{tcolorbox}[width=1\textwidth, colback=lightblue, title={\textbf{Correction critique prompt.}}]

    You are a professional expert in critique who can evaluate the responses generated by large multimodal models and analyze response quality. \\

    \textbf{\#}: The following are the \texttt{question} and the generated \texttt{response} from large multimodal models.  \\
    \textbf{------------------------------------------------------------------------------------------------------------------------}

    --- Start of Question ---\\
    \textcolor{ora}{\$Question (include images)\$}\\
    
    --- End of Question---\\

     --- Start of Response  ---\\
    \textcolor{ora}{\$Response \$}\\
    --- End of Response ---\\

   \textbf{------------------------------------------------------------------------------------------------------------------------}\\

    \textbf{\#Very important rules !!!}:\\
    \textbf{\#\#}1. Carefully understand the question, and then analyze the response. \\
    \textbf{\#\#}2. The original response is not good enough, and you should give your own response to better answer the question.

    \textbf{\#Output Format}:\\
    Provide the evaluation in JSON format as follows:\\
    \verb|```|json\\
    \{\\
        \quad "modified answer": "Since the response is unsatisfactory, give your own response here, due to the question",\\      
    \}\\
    \verb|```|\\

    \end{tcolorbox}
\end{center}
\caption{Correction critique prompt.} \label{ap:prompt_correlation}
\end{figure*}


\begin{figure*}[h!]
\begin{center}
    \fontsize{8.4}{8.4} \selectfont
    \begin{tcolorbox}[width=1\textwidth, colback=lightblue, title={\textbf{Comparative critique prompt.}}]

    You are a professional comparative critique evaluation judge who can evaluate the responses generated by two different large multimodal models and choose the better one. \\

    \textbf{\#}: The following are the \texttt{question} and two generated \texttt{response} from two different models.  \\
    \textbf{------------------------------------------------------------------------------------------------------------------------}

    --- Start of Question ---\\
    \textcolor{ora}{\$Question (include images)\$}\\
    
    --- End of Question---\\

     --- Start of Response A ---\\
    \textcolor{ora}{\$Response A\$}\\
    --- End of Response A---\\

    --- Start of Response B ---\\
    \textcolor{ora}{\$Response B\$}\\
    --- End of Response B---

   \textbf{------------------------------------------------------------------------------------------------------------------------}\\

    \textbf{\#Very important rules !!!}:\\
    \textbf{\#\#}1. Carefully compare the two responses, and then choose the better one. \\
    \textbf{\#\#}2. Please first generate your explanation for the choice by comparing the two responses, and then you must clearly state your choice following the format: "choice": X, where X is A or B, corresponding to response A and response B.

    \textbf{\#Output Format}:\\
    Provide the evaluation in JSON format as follows:\\
    \verb|```|json\\
    \{\\
        \quad"choice": "chose the better response quality model and indicate your choice is ["A","B"]",\\
        \quad"explanation": "Based on the two responses, give explanation about the choice given by yourself", \\
    \}\\
    \verb|```|\\

    \end{tcolorbox}
\end{center}
\caption{Comparative critique prompt.} \label{ap:prompt_comparative}
\end{figure*}

\clearpage
\section{Experimental Results}
\label{Apx:c}

\begin{table*}[h]
\centering

\resizebox{\textwidth}{!}{

\begin{tabular}{@{}lccccccccccc}
\toprule
\multirow{2}{*}{\textbf{Model}} 
& \multicolumn{9}{c}{\textbf{Task types}} 
\\
\cline{2-9} 
& \textbf{Perception } & \textbf{Planning} & \textbf{Knowledge } &\textbf{Information Extraction} & \textbf{Mathematics} & \textbf{Coding} & \textbf{Science} & \textbf{Metric} & \textbf{Avg.} \\
\midrule

\rowcolor{pink!50}
\multicolumn{10}{c}{\textit{Proprietary Models}}\\
\textbf{o4-mini} & \multicolumn{1}{c}{7.913} & 7.987
& \multicolumn{1}{c}{7.817} & 7.806
& \multicolumn{1}{c}{8.080} & 80.84
& \multicolumn{1}{c}{7.869} & 8.049
& \multicolumn{1}{c}{7.924} \\
\textbf{GPT-4o} & \multicolumn{1}{c}{7.537} & 7.473
& \multicolumn{1}{c}{7.288} & 7.496
& \multicolumn{1}{c}{7.444} & 7.523
& \multicolumn{1}{c}{7.686} & 7.937
& \multicolumn{1}{c}{7.499} \\
\textbf{GPT-4o-mini} & \multicolumn{1}{c}{6.65} & 6.653
& \multicolumn{1}{c}{6.691} & 6.831
& \multicolumn{1}{c}{6.346} & 6.437
& \multicolumn{1}{c}{6.516} & 6.742
& \multicolumn{1}{c}{6.634} \\
\textbf{Claude-3.7-sonnet} & \multicolumn{1}{c}{8.084} & 8.066
& \multicolumn{1}{c}{8.137} & 8.176
& \multicolumn{1}{c}{8.283} & 8.058
& \multicolumn{1}{c}{8.230} & 7.178
& \multicolumn{1}{c}{8.113} \\
\textbf{Gemini-2.5-flash} & \multicolumn{1}{c}{6.478} & 6.358
& \multicolumn{1}{c}{6.242} & 6.386
& \multicolumn{1}{c}{6.596} & 6.786
& \multicolumn{1}{c}{7.121} & 6.784
& \multicolumn{1}{c}{6.495} \\
\textbf{Gemini-2.5-pro} & \multicolumn{1}{>{\columncolor{gray!20}}c}{8.524} & \cellcolor{gray!20}8.831
& \multicolumn{1}{>{\columncolor{gray!20}}c}{8.380} & \cellcolor{gray!20}8.325
& \multicolumn{1}{>{\columncolor{gray!20}}c}{8.987} & \cellcolor{gray!20}8.524
& \multicolumn{1}{>{\columncolor{gray!20}}c}{8.537} & \cellcolor{gray!20}8.786
& \multicolumn{1}{>{\columncolor{gray!40}}c}{8.558} \\
\textbf{Grok-2-vision} & \multicolumn{1}{c}{7.567} & 7.654
& \multicolumn{1}{c}{7.509} & 7.112
& \multicolumn{1}{c}{7.474} & 7.645
& \multicolumn{1}{c}{7.760} & 7.695
& \multicolumn{1}{c}{7.523}\\
\midrule
\rowcolor{green!30}
\multicolumn{10}{c}{\textit{Open-weight Models (Larger than 30B)}}\\
\textbf{Qwen2.5-vl-32b} & \multicolumn{1}{c}{8.245} & 8.286
& \multicolumn{1}{c}{7.983} & 8.237
& \multicolumn{1}{c}{7.861} & 8.413
& \multicolumn{1}{c}{8.377} & 8.566
& \multicolumn{1}{c}{8.208} \\

\textbf{Qwen2.5-vl-72b} & \multicolumn{1}{c}{6.933} & 7.332
& \multicolumn{1}{c}{6.892} & 6.194
& \multicolumn{1}{c}{6.725} & 7.430
& \multicolumn{1}{c}{7.155} & 6.914
& \multicolumn{1}{c}{6.931} \\
\textbf{Pixtral-large}& \multicolumn{1}{c}{7.445} & 7.889
& \multicolumn{1}{c}{7.289} & 7.495
& \multicolumn{1}{c}{7.272} & 7.459
& \multicolumn{1}{c}{7.842} & 7.263
& \multicolumn{1}{c}{7.489}\\
\textbf{Llama-4-maverick}& \multicolumn{1}{c}{5.785} & 5.603
& \multicolumn{1}{c}{5.648} & 6.406
& \multicolumn{1}{c}{5.484} & 5.826
& \multicolumn{1}{c}{6.054} & 6.217
& \multicolumn{1}{c}{5.811}\\
\midrule
\rowcolor{green!30}
\multicolumn{10}{c}{\textit{Open-weight Models (Less than 30B)}}\\
\textbf{Gemma-3-4b} & \multicolumn{1}{c}{5.245} & 5.211
& \multicolumn{1}{c}{5.239} & 4.663
& \multicolumn{1}{c}{4.962} & 4.680
& \multicolumn{1}{c}{5.049} & 6.078
& \multicolumn{1}{c}{5.130} \\
\textbf{Qwen2.5-vl-7b} & \multicolumn{1}{c}{4.509} & 4.889
& \multicolumn{1}{c}{4.300} & 4.854
& \multicolumn{1}{c}{4.847} & 4.658
& \multicolumn{1}{c}{4.642} & 4.575
& \multicolumn{1}{c}{4.617} \\
\textbf{Llama-3.2-11b-vision} & \multicolumn{1}{c}{5.230} & 5.266
& \multicolumn{1}{c}{5.233} & 4.876
& \multicolumn{1}{c}{5.162} & 4.927
& \multicolumn{1}{c}{5.198} & 5.791
& \multicolumn{1}{c}{5.185} \\
\textbf{Pixtral-12b}& \multicolumn{1}{c}{5.317} & 5.482
& \multicolumn{1}{c}{5.218} & 4.623
& \multicolumn{1}{c}{4.899} & 5.034
& \multicolumn{1}{c}{5.599} & 5.139
& \multicolumn{1}{c}{5.201}\\
\textbf{Gemma-3-12b} & \multicolumn{1}{c}{6.572} & 6.571
& \multicolumn{1}{c}{6.761} & 5.678
& \multicolumn{1}{c}{6.463} & 6.838
& \multicolumn{1}{c}{6.992} & 6.765
& \multicolumn{1}{c}{6.566} \\
\textbf{Gemma-3-27b} & \multicolumn{1}{c}{7.285} & 7.235
& \multicolumn{1}{c}{7.032} & 6.135
& \multicolumn{1}{c}{7.169} & 7.262
& \multicolumn{1}{c}{7.290} & 7.031
& \multicolumn{1}{c}{7.107} \\
\textbf{Llama-4-scout} & \multicolumn{1}{c}{5.996} & 5.714
& \multicolumn{1}{c}{5.723} & 5.563
& \multicolumn{1}{c}{5.543} & 5.537
& \multicolumn{1}{c}{5.985} & 5.633
& \multicolumn{1}{c}{5.771}\\

\bottomrule
\end{tabular}
}
\caption{Critique scores of \textbf{{Core}} subset on different task types.}\label{tab:core_socre}
\end{table*}

\begin{table*}[h]
\centering

\resizebox{\textwidth}{!}{

\begin{tabular}{@{}lccccccccccc}
\toprule
\multirow{2}{*}{\textbf{Model}} 
& \multicolumn{9}{c}{\textbf{Task types}} 
\\
\cline{2-9} 
& \textbf{Perception } & \textbf{Planning} & \textbf{Knowledge } &\textbf{Information Extraction} & \textbf{Mathematics} & \textbf{Coding} & \textbf{Science} & \textbf{Metric} & \textbf{Avg.} \\
\midrule

\rowcolor{pink!50}
\multicolumn{10}{c}{\textit{Proprietary Models}}\\
\textbf{o4-mini} & \multicolumn{1}{c}{7.950} & 8.004
& \multicolumn{1}{c}{7.909} & 7.934
& \multicolumn{1}{c}{8.05} & 7.790
& \multicolumn{1}{c}{8.103} & 7.111
& \multicolumn{1}{c}{7.953} \\
\textbf{GPT-4o} & \multicolumn{1}{c}{7.285} & 7.455
& \multicolumn{1}{c}{7.596} & 7.548
& \multicolumn{1}{c}{7.370} & 7.418
& \multicolumn{1}{c}{7.180} & 7.418
& \multicolumn{1}{c}{7.429} \\
\textbf{GPT-4o-mini} & \multicolumn{1}{c}{6.508} & 6.680
& \multicolumn{1}{c}{6.592} & 6.507
& \multicolumn{1}{c}{6.469} & 6.457
& \multicolumn{1}{c}{6.398} & 6.167
& \multicolumn{1}{c}{6.534} \\
\textbf{Claude-3.7-sonnet} & \multicolumn{1}{c}{8.102} & 8.158
& \multicolumn{1}{c}{8.070} & 7.961
& \multicolumn{1}{c}{8.154} & 8.104
& \multicolumn{1}{c}{8.133} & 6.944
& \multicolumn{1}{c}{8.080} \\
\textbf{Gemini-2.5-flash} & \multicolumn{1}{c}{6.384} & 6.459
& \multicolumn{1}{c}{6.225} & 6.418
& \multicolumn{1}{c}{6.768} & 6.623
& \multicolumn{1}{c}{7.010} & 6.722
& \multicolumn{1}{c}{6.460} \\
\textbf{Gemini-2.5-pro} & \multicolumn{1}{>{\columncolor{gray!20}}c}{8.542} & \cellcolor{gray!20}8.692
& \multicolumn{1}{>{\columncolor{gray!20}}c}{8.377} & \cellcolor{gray!20}8.462
& \multicolumn{1}{>{\columncolor{gray!20}}c}{8.933} & \cellcolor{gray!20}8.427
& \multicolumn{1}{>{\columncolor{gray!20}}c}{8.500} & 8.750
& \multicolumn{1}{>{\columncolor{gray!40}}c}{8.549} \\
\textbf{Grok-2-vision} & \multicolumn{1}{c}{7.476} & 7.861
& \multicolumn{1}{c}{7.606} & 6.811
& \multicolumn{1}{c}{7.798} & 7.250
& \multicolumn{1}{c}{7.765} & 6.944
& \multicolumn{1}{c}{7.490}\\
\midrule
\rowcolor{green!30}
\multicolumn{10}{c}{\textit{Open-weight Models (Larger than 30B)}}\\
\textbf{Qwen2.5-vl-32b} & \multicolumn{1}{c}{7.946} & 8.496
& \multicolumn{1}{c}{8.318} & 8.007
& \multicolumn{1}{c}{7.844} & 8.074
& \multicolumn{1}{c}{8.329} & \cellcolor{gray!20}9.000
& \multicolumn{1}{c}{8.138} \\
\textbf{Qwen2.5-vl-72b} & \multicolumn{1}{c}{7.180} & 7.149
& \multicolumn{1}{c}{7.270} & 7.090
& \multicolumn{1}{c}{6.953} & 7.200
& \multicolumn{1}{c}{7.330} & 6.176
& \multicolumn{1}{c}{7.157} \\

\textbf{Pixtral-large} & \multicolumn{1}{c}{7.480} & 7.788
& \multicolumn{1}{c}{7.497} & 7.254
& \multicolumn{1}{c}{7.588} & 7.583
& \multicolumn{1}{c}{7.800} & 7.000
& \multicolumn{1}{c}{7.531}\\

\textbf{Llama-4-maverick} & \multicolumn{1}{c}{6.151} & 5.672
& \multicolumn{1}{c}{5.897} & 6.232
& \multicolumn{1}{c}{5.804} & 5.808
& \multicolumn{1}{c}{6.102} & 5.500
& \multicolumn{1}{c}{5.971}\\
\midrule
\rowcolor{green!30}
\multicolumn{10}{c}{\textit{Open-weight Models (Less than 30B)}}\\
\textbf{Gemma-3-4b} & \multicolumn{1}{c}{5.670} & 5.697
& \multicolumn{1}{c}{5.511} & 4.706
& \multicolumn{1}{c}{5.658} & 5.211
& \multicolumn{1}{c}{5.769} & 5.533
& \multicolumn{1}{c}{5.509} \\
\textbf{Qwen2.5-vl-7b} & \multicolumn{1}{c}{5.113} & 5.391
& \multicolumn{1}{c}{5.280} & 4.453
& \multicolumn{1}{c}{4.925} & 5.078
& \multicolumn{1}{c}{5.122} & 5.846
& \multicolumn{1}{c}{5.093} \\

\textbf{Llama-3.2-11b-vision} & \multicolumn{1}{c}{5.113} & 5.391
& \multicolumn{1}{c}{5.280} & 4.453
& \multicolumn{1}{c}{4.925} & 5.078
& \multicolumn{1}{c}{5.122} & 5.846
& \multicolumn{1}{c}{5.093} \\
\textbf{Pixtral-12b} & \multicolumn{1}{c}{0.657} & 0.833
& \multicolumn{1}{c}{0.668} & 0.621
& \multicolumn{1}{c}{0.752} & 0.620
& \multicolumn{1}{c}{0.689} & 0.444
& \multicolumn{1}{c}{0.687}\\
\textbf{Gemma-3-12b} & \multicolumn{1}{c}{6.429} & 6.526
& \multicolumn{1}{c}{6.573} & 6.000
& \multicolumn{1}{c}{6.361} & 6.268
& \multicolumn{1}{c}{6.863} & 5.389
& \multicolumn{1}{c}{6.419} \\
\textbf{Gemma-3-27b} & \multicolumn{1}{c}{6.987} & 7.009
& \multicolumn{1}{c}{7.173} & 5.987
& \multicolumn{1}{c}{7.270} & 6.194
& \multicolumn{1}{c}{7.333} & 5.750
& \multicolumn{1}{c}{6.921} \\
\textbf{Llama-4-scout} & \multicolumn{1}{c}{6.087} & 5.713
& \multicolumn{1}{c}{5.891} & 5.652
& \multicolumn{1}{c}{5.446} & 5.225
& \multicolumn{1}{c}{6.172} & 5.667
& \multicolumn{1}{c}{5.822}\\

\bottomrule
\end{tabular}
}
\caption{Critique scores of \textbf{{Core-single-image}} subset on different task types.}\label{tab:core-single-score}
\end{table*}


\begin{table*}[h]
\centering

\resizebox{\textwidth}{!}{

\begin{tabular}{@{}lccccccccccc}
\toprule
\multirow{2}{*}{\textbf{Model}} 
& \multicolumn{9}{c}{\textbf{Task types}} 
\\
\cline{2-9} 
& \textbf{Perception } & \textbf{Planning} & \textbf{Knowledge } &\textbf{Information Extraction} & \textbf{Mathematics} & \textbf{Coding} & \textbf{Science} & \textbf{Metric} & \textbf{Avg.} \\
\midrule

\rowcolor{pink!50}
\multicolumn{10}{c}{\textit{Proprietary Models}}\\
\textbf{o4-mini} & \multicolumn{1}{c}{7.817} & 8.077
& \multicolumn{1}{c}{7.953} & 7.510
& \multicolumn{1}{c}{-} & 7.727
& \multicolumn{1}{c}{8.500} & 7.857
& \multicolumn{1}{c}{7.877} \\
\textbf{GPT-4o} & \multicolumn{1}{c}{7.808} & 7.933
& \multicolumn{1}{c}{7.818} & 7.980
& \multicolumn{1}{c}{-} & 7.700
& \multicolumn{1}{c}{7.818} & 5.600
& \multicolumn{1}{c}{7.807} \\
\textbf{GPT-4o-mini} & \multicolumn{1}{c}{6.869} & 6.615
& \multicolumn{1}{c}{6.747} & 5.745
& \multicolumn{1}{c}{-} & 6.833
& \multicolumn{1}{c}{5.833} & 5.857
& \multicolumn{1}{c}{6.549} \\
\textbf{Claude-3.7-sonnet} & \multicolumn{1}{c}{8.233} & 8.591
& \multicolumn{1}{c}{8.169} & 7.627
& \multicolumn{1}{c}{-} & 8.636
& \multicolumn{1}{c}{7.818} & 7.00
& \multicolumn{1}{c}{8.102} \\
\textbf{Gemini-2.5-flash} & \multicolumn{1}{c}{6.049} & 6.615
& \multicolumn{1}{c}{6.718} & 6.549
& \multicolumn{1}{c}{-} & 6.083
& \multicolumn{1}{c}{6.500} & 5.714
& \multicolumn{1}{c}{6.500} \\
\textbf{Gemini-2.5-pro} & \multicolumn{1}{c}{8.327} & 8.875
& \multicolumn{1}{c}{8.340} & 7.878
& \multicolumn{1}{c}{-} & 8.273
& \multicolumn{1}{c}{7.182} & 6.000
& \multicolumn{1}{c}{8.246} \\
\textbf{Grok-2-vision}& \multicolumn{1}{c}{8.117} & 8.417
& \multicolumn{1}{c}{8.242} & 7.314
& \multicolumn{1}{c}{-} & 8.500
& \multicolumn{1}{c}{8.333} & 7.500
& \multicolumn{1}{c}{8.066} \\
\midrule
\rowcolor{green!30}
\multicolumn{10}{c}{\textit{Open-weight Models (Larger than 30B)}}\\
\textbf{Qwen2.5-vl-32b} & \multicolumn{1}{>{\columncolor{gray!20}}c}{8.882} & \cellcolor{gray!20}9.000
& \multicolumn{1}{>{\columncolor{gray!20}}c}{8.419} & \cellcolor{gray!20}8.033
& \multicolumn{1}{c}{-} & \cellcolor{gray!20}9.300
& \multicolumn{1}{>{\columncolor{gray!20}}c}{9.364} & \cellcolor{gray!20}7.800
& \multicolumn{1}{>{\columncolor{gray!40}}c}{8.566} \\

\textbf{Qwen2.5-vl-72b} & \multicolumn{1}{c}{7.000} & 7.042
& \multicolumn{1}{c}{7.034} & 6.800
& \multicolumn{1}{c}{-} & 7.000
& \multicolumn{1}{c}{7.500} & 5.429
& \multicolumn{1}{c}{6.974} \\

\textbf{Pixtral-large}& \multicolumn{1}{c}{7.855} & 7.895
& \multicolumn{1}{c}{7.832} & 7.575
& \multicolumn{1}{c}{-} & 7.818
& \multicolumn{1}{c}{7.833} & 5.000
& \multicolumn{1}{c}{7.743}\\
\textbf{Llama-4-maverick}& \multicolumn{1}{c}{6.441} & 6.077
& \multicolumn{1}{c}{6.541} & 6.040
& \multicolumn{1}{c}{-} & 6.000
& \multicolumn{1}{c}{6.167} & 5.000
& \multicolumn{1}{c}{6.342}\\
\midrule
\rowcolor{green!30}
\multicolumn{10}{c}{\textit{Open-weight Models (Less than 30B)}}\\
\textbf{Gemma-3-4b} & \multicolumn{1}{c}{5.696} & 7.214
& \multicolumn{1}{c}{6.638} & 5.737
& \multicolumn{1}{c}{-} & 2.583
& \multicolumn{1}{c}{5.000} & 3.000
& \multicolumn{1}{c}{6.155} \\
\textbf{Qwen2.5-vl-7b} & \multicolumn{1}{c}{0.796} & 0.727
& \multicolumn{1}{c}{0.682} & 0.722
& \multicolumn{1}{c}{-} & 0.750
& \multicolumn{1}{c}{0.600} & 0.0
& \multicolumn{1}{c}{0.711} \\

\textbf{Llama-3.2-11b-vision} & \multicolumn{1}{c}{5.230} & 5.640
& \multicolumn{1}{c}{5.333} & 4.941
& \multicolumn{1}{c}{-} & 3.250
& \multicolumn{1}{c}{4.750} & 5.714
& \multicolumn{1}{c}{5.179} \\
\textbf{Pixtral-12b}& \multicolumn{1}{c}{5.818} & 5.417
& \multicolumn{1}{c}{5.987} & 5.061
& \multicolumn{1}{c}{-} & 5.250
& \multicolumn{1}{c}{4.000} & 3.667
& \multicolumn{1}{c}{5.700}\\
\textbf{Gemma-3-12b} & \multicolumn{1}{c}{6.700} & -
& \multicolumn{1}{c}{7.025} & 5.176
& \multicolumn{1}{c}{-} & 5.889
& \multicolumn{1}{c}{7.600} & 4.000
& \multicolumn{1}{c}{6.744} \\
\textbf{Gemma-3-27b} & \multicolumn{1}{c}{6.467} & 8.474
& \multicolumn{1}{c}{7.500} & 5.889
& \multicolumn{1}{c}{-} & 6.875
& \multicolumn{1}{c}{7.417} & 4.000
& \multicolumn{1}{c}{7.297} \\
\textbf{Llama-4-scout} & \multicolumn{1}{c}{6.500} & 6.318
& \multicolumn{1}{c}{6.386} & 5.810
& \multicolumn{1}{c}{-} & 6.083
& \multicolumn{1}{c}{5.917} & 9.000
& \multicolumn{1}{c}{6.336}\\


\bottomrule
\end{tabular}
}
\caption{Critique scores of \textbf{{Open}} subset on different task types.}\label{tab:open-socre}
\end{table*}

\begin{table*}[h]
\centering

\resizebox{\textwidth}{!}{

\begin{tabular}{@{}lccccccccccc}
\toprule
\multirow{2}{*}{\textbf{Model}} 
& \multicolumn{9}{c}{\textbf{Task types}} 
\\
\cline{2-9} 
& \textbf{Perception } & \textbf{Planning} & \textbf{Knowledge } &\textbf{Information Extraction} & \textbf{Mathematics} & \textbf{Coding} & \textbf{Science} & \textbf{Metric} & \textbf{Avg.} \\
\midrule

\rowcolor{pink!50}
\multicolumn{10}{c}{\textit{Proprietary Models}}\\
\textbf{o4-mini} & \multicolumn{1}{c}{8.037} & 8.087
& \multicolumn{1}{c}{7.897} & 8.083
& \multicolumn{1}{c}{-} & 8.083
& \multicolumn{1}{c}{8.167} & -
& \multicolumn{1}{c}{7.976} \\
\textbf{GPT-4o} & \multicolumn{1}{c}{7.316} & 7.875
& \multicolumn{1}{c}{7.575} & 8.500
& \multicolumn{1}{c}{-} & 7.500
& \multicolumn{1}{>{\columncolor{gray!20}}c}{8.556} & -
& \multicolumn{1}{c}{7.637} \\
\textbf{GPT-4o-mini} & \multicolumn{1}{c}{0.593} & 0.739
& \multicolumn{1}{c}{0.679} & 0.750
& \multicolumn{1}{c}{-} & 0.917
& \multicolumn{1}{c}{0.900} & -
& \multicolumn{1}{c}{0.690} \\
\textbf{Claude-3.7-sonnet} & \multicolumn{1}{c}{8.038} & 8.684
& \multicolumn{1}{c}{7.916} & 8.667
& \multicolumn{1}{c}{-} & 8.636
& \multicolumn{1}{c}{8.333} & -
& \multicolumn{1}{c}{8.097} \\
\textbf{Gemini-2.5-flash} & \multicolumn{1}{c}{6.264} & 6.238
& \multicolumn{1}{c}{6.276} & 6.833
& \multicolumn{1}{c}{-} & 6.583
& \multicolumn{1}{c}{6.833} & -
& \multicolumn{1}{c}{6.340} \\
\textbf{Gemini-2.5-pro} & \multicolumn{1}{c}{8.431} & \cellcolor{gray!20}8.895
& \multicolumn{1}{c}{8.283} & 8.750
& \multicolumn{1}{c}{-} & 8.091
& \multicolumn{1}{c}{7.667} & -
& \multicolumn{1}{c}{8.325} \\
\textbf{Grok-2-vision}& \multicolumn{1}{c}{8.260} & 8.261
& \multicolumn{1}{c}{8.235} & 8.750
& \multicolumn{1}{c}{-} & 8.500
& \multicolumn{1}{c}{8.143} & -
& \multicolumn{1}{c}{8.274}\\
\midrule
\rowcolor{green!30}
\multicolumn{10}{c}{\textit{Open-weight Models (Larger than 30B)}}\\
\textbf{Qwen2.5-vl-32b} & \multicolumn{1}{>{\columncolor{gray!20}}c}{8.857} & 8.667
& \multicolumn{1}{>{\columncolor{gray!20}}c}{8.392} & \cellcolor{gray!20}9.000
& \multicolumn{1}{c}{-} & \cellcolor{gray!20}8.667
& \multicolumn{1}{c}{-} & -
& \multicolumn{1}{>{\columncolor{gray!40}}c}{8.495} \\
\textbf{Qwen2.5-vl-72b} & \multicolumn{1}{c}{6.843} & 7.909
& \multicolumn{1}{c}{6.788} & 7.667
& \multicolumn{1}{c}{-} & 7.333
& \multicolumn{1}{c}{7.167} & -
& \multicolumn{1}{c}{6.992} \\
\textbf{Pixtral-large} & \multicolumn{1}{c}{7.794} & 8.300
& \multicolumn{1}{c}{7.739} & 8.714
& \multicolumn{1}{c}{-} & 7.636
& \multicolumn{1}{c}{7.333} & -
& \multicolumn{1}{c}{7.784}\\
\textbf{Llama-4-maverick} & \multicolumn{1}{c}{6.333} & 5.227
& \multicolumn{1}{c}{6.447} & 6.500
& \multicolumn{1}{c}{-} & 6.083
& \multicolumn{1}{c}{5.455} & -
& \multicolumn{1}{c}{6.250}\\
\midrule
\rowcolor{green!30}
\multicolumn{10}{c}{\textit{Open-weight Models (Less than 30B)}}\\
\textbf{Gemma-3-4b} & \multicolumn{1}{c}{7.042} & 7.313
& \multicolumn{1}{c}{6.828} & 5.917
& \multicolumn{1}{c}{-} & 3.833
& \multicolumn{1}{c}{4.667} & -
& \multicolumn{1}{c}{6.643} \\

\textbf{Qwen2.5-vl-7b} & \multicolumn{1}{c}{4.721} & 4.905
& \multicolumn{1}{c}{4.750} & 5.222
& \multicolumn{1}{c}{-} & 5.273
& \multicolumn{1}{c}{4.727} & -
& \multicolumn{1}{c}{4.806} \\
\textbf{Llama-3.2-11b-vision} & \multicolumn{1}{c}{5.143} & 5.222
& \multicolumn{1}{c}{5.450} & 6.091
& \multicolumn{1}{c}{-} & 4.778
& \multicolumn{1}{c}{5.091} & -
& \multicolumn{1}{c}{5.351} \\
\textbf{Pixtral-12b}& \multicolumn{1}{c}{5.755} & 5.130
& \multicolumn{1}{c}{5.815} & 6.417
& \multicolumn{1}{c}{-} & 5.333
& \multicolumn{1}{c}{6.200} & -
& \multicolumn{1}{c}{5.759} \\
\textbf{Gemma-3-12b} & \multicolumn{1}{c}{6.800} & 6.250
& \multicolumn{1}{c}{7.102} & 6.444
& \multicolumn{1}{c}{-} & 6.400
& \multicolumn{1}{c}{7.091} & -
& \multicolumn{1}{c}{ 6.944} \\
\textbf{Gemma-3-27b} & \multicolumn{1}{c}{7.489} & 7.957
& \multicolumn{1}{c}{7.872} & 7.667
& \multicolumn{1}{c}{-} & 6.818
& \multicolumn{1}{c}{6.909} & -
& \multicolumn{1}{c}{7.700} \\
\textbf{Llama-4-scout}& \multicolumn{1}{c}{6.620} & 6.095
& \multicolumn{1}{c}{6.535} & 7.250
& \multicolumn{1}{c}{-} & 6.583
& \multicolumn{1}{c}{5.800} & -
& \multicolumn{1}{c}{6.521}  \\

\bottomrule
\end{tabular}
}
\caption{Critique scores of \textbf{{Open-singel-image}} subset on different task types.}\label{tab:open-single-score}
\end{table*}

\begin{table*}[h]
\centering

\resizebox{\textwidth}{!}{

\begin{tabular}{@{}lccccccccccc}
\toprule
\multirow{2}{*}{\textbf{Model}} 
& \multicolumn{9}{c}{\textbf{Task types}} 
\\
\cline{2-9} 
& \textbf{Perception } & \textbf{Planning} & \textbf{Knowledge } &\textbf{Information Extraction} & \textbf{Mathematics} & \textbf{Coding} & \textbf{Science} & \textbf{Metric} & \textbf{Avg.} \\
\midrule

\rowcolor{pink!50}
\multicolumn{10}{c}{\textit{Proprietary Models}}\\
\textbf{o4-mini} & \multicolumn{1}{>{\columncolor{gray!20}}c}{0.872} & \cellcolor{gray!20}0.974
& \multicolumn{1}{>{\columncolor{gray!20}}c}{0.868} & 0.940
& \multicolumn{1}{>{\columncolor{gray!20}}c}{0.899} & \cellcolor{gray!20}0.916
& \multicolumn{1}{>{\columncolor{gray!20}}c}{0.869} & \cellcolor{gray!20}0.783
& \multicolumn{1}{>{\columncolor{gray!40}}c}{0.896} \\
\textbf{GPT-4o} & \multicolumn{1}{c}{0.847} & 0.917
& \multicolumn{1}{c}{0.809} & 0.848
& \multicolumn{1}{c}{0.792} & 0.754
& \multicolumn{1}{c}{0.821} & 0.738
& \multicolumn{1}{c}{0.832} \\
\textbf{GPT-4o-mini} & \multicolumn{1}{c}{0.818} & 0.852
& \multicolumn{1}{c}{0.821} & 0.852
& \multicolumn{1}{c}{0.840} & 0.821
& \multicolumn{1}{c}{0.795} & 0.678
& \multicolumn{1}{c}{0.833} \\
\textbf{Claude-3.7-sonnet} & \multicolumn{1}{c}{0.808} & 0.925
& \multicolumn{1}{c}{0.821} & 0.884
& \multicolumn{1}{c}{0.870} & 0.844
& \multicolumn{1}{c}{0.814} & 0.698
& \multicolumn{1}{c}{0.840} \\
\textbf{Gemini-2.5-flash} & \multicolumn{1}{c}{0.840} & 0.903
& \multicolumn{1}{c}{0.772} & 0.841
& \multicolumn{1}{c}{0.831} & 0.818
& \multicolumn{1}{c}{0.814} & 0.667
& \multicolumn{1}{c}{0.826} \\
\textbf{Gemini-2.5-pro} & \multicolumn{1}{c}{0.826} & 0.941
& \multicolumn{1}{c}{0.838} & \cellcolor{gray!20}0.943
& \multicolumn{1}{c}{0.898} & 0.889
& \multicolumn{1}{c}{0.789} & 0.775
& \multicolumn{1}{c}{0.865} \\
\textbf{Grok-2-vision}& \multicolumn{1}{c}{0.782} & 0.875
& \multicolumn{1}{c}{0.782} & 0.841
& \multicolumn{1}{c}{0.847} & 0.794
& \multicolumn{1}{c}{0.769} & 0.671
& \multicolumn{1}{c}{0.803}\\
\midrule
\rowcolor{green!30}
\multicolumn{10}{c}{\textit{Open-weight Models (Larger than 30B)}}\\
\textbf{Qwen2.5-vl-32b} & \multicolumn{1}{c}{0.823} & 0.923
& \multicolumn{1}{c}{0.797} & 0.882
& \multicolumn{1}{c}{0.821} & 0.860
& \multicolumn{1}{c}{0.831} & 0.738
& \multicolumn{1}{c}{0.839} \\

\textbf{Qwen2.5-vl-72b} & \multicolumn{1}{c}{0.805} & 0.935
& \multicolumn{1}{c}{0.802} & 0.852
& \multicolumn{1}{c}{0.790} & 0.845
& \multicolumn{1}{c}{0.809} & 0.667
& \multicolumn{1}{c}{0.824} \\

\textbf{Pixtral-large}& \multicolumn{1}{c}{0.823} & 0.945
& \multicolumn{1}{c}{0.774} & 0.881
& \multicolumn{1}{c}{0.810} & 0.819
& \multicolumn{1}{c}{0.815} & 0.675
& \multicolumn{1}{c}{0.828} \\
\textbf{Llama-4-maverick}& \multicolumn{1}{c}{0.719} & 0.867
& \multicolumn{1}{c}{0.727} & 0.787
& \multicolumn{1}{c}{0.747} & 0.710
& \multicolumn{1}{c}{0.748} & 0.586
& \multicolumn{1}{c}{ 0.748} \\
\midrule
\rowcolor{green!30}
\multicolumn{10}{c}{\textit{Open-weight Models (Less than 30B)}}\\
\textbf{Gemma-3-4b} & \multicolumn{1}{c}{0.490} & 0.621
& \multicolumn{1}{c}{0.501} & 0.462
& \multicolumn{1}{c}{0.535} & 0.426
& \multicolumn{1}{c}{0.490} & 0.469
& \multicolumn{1}{c}{0.508} \\

\textbf{Qwen2.5-vl-7b} & \multicolumn{1}{c}{0.736} & 0.893
& \multicolumn{1}{c}{0.747} & 0.805
& \multicolumn{1}{c}{0.800} & 0.804
& \multicolumn{1}{c}{0.796} & 0.716
& \multicolumn{1}{c}{0.783} \\

\textbf{Llama-3.2-11b-vision} & \multicolumn{1}{c}{0.693} & 0.832
& \multicolumn{1}{c}{0.703} & 0.724
& \multicolumn{1}{c}{0.789} & 0.793
& \multicolumn{1}{c}{0.610} & 0.618
& \multicolumn{1}{c}{0.721} \\
\textbf{Pixtral-12b}& \multicolumn{1}{c}{0.707} & 0.830
& \multicolumn{1}{c}{0.688} & 0.584
& \multicolumn{1}{c}{0.704} & 0.649
& \multicolumn{1}{c}{0.796} & 0.588
& \multicolumn{1}{c}{0.704} \\
\textbf{Gemma-3-12b} & \multicolumn{1}{c}{0.744} & 0.877
& \multicolumn{1}{c}{0.765} & 0.661
& \multicolumn{1}{c}{0.769} & 0.757
& \multicolumn{1}{c}{0.768} & 0.638
& \multicolumn{1}{c}{0.759} \\
\textbf{Gemma-3-27b} & \multicolumn{1}{c}{0.816} & 0.922
& \multicolumn{1}{c}{0.778} & 0.684
& \multicolumn{1}{c}{0.809} & 0.824
& \multicolumn{1}{c}{0.739} & 0.727
& \multicolumn{1}{c}{0.804} \\
\textbf{Llama-4-scout} & \multicolumn{1}{c}{0.748} & 0.890
& \multicolumn{1}{c}{0.745} & 0.693
& \multicolumn{1}{c}{0.739} & 0.724
& \multicolumn{1}{c}{0.752} & 0.656
& \multicolumn{1}{c}{ 0.757} \\

\bottomrule
\end{tabular}
}
\caption{ $\mathrm{ACC_{critic}}$ of \textbf{{Core}} subset on different task types.}\label{tab:core-acc}
\end{table*}

\begin{table*}[h]
\centering

\resizebox{\textwidth}{!}{

\begin{tabular}{@{}lccccccccccc}
\toprule
\multirow{2}{*}{\textbf{Model}} 
& \multicolumn{9}{c}{\textbf{Task types}} 
\\
\cline{2-9} 
& \textbf{Perception } & \textbf{Planning} & \textbf{Knowledge } &\textbf{Information Extraction} & \textbf{Mathematics} & \textbf{Coding} & \textbf{Science} & \textbf{Metric} & \textbf{Avg.} \\
\midrule

\rowcolor{pink!50}
\multicolumn{10}{c}{\textit{Proprietary Models}}\\
\textbf{o4-mini} & \multicolumn{1}{>{\columncolor{gray!20}}c}{0.843} & \cellcolor{gray!20}0.975
& \multicolumn{1}{>{\columncolor{gray!20}}c}{0.868} & 0.938
& \multicolumn{1}{>{\columncolor{gray!20}}c}{0.916} & \cellcolor{gray!20}0.951
& \multicolumn{1}{>{\columncolor{gray!20}}c}{0.880} & 0.889
& \multicolumn{1}{>{\columncolor{gray!40}}c}{0.897} \\
\textbf{GPT-4o} & \multicolumn{1}{c}{0.811} & 0.916
& \multicolumn{1}{c}{0.805} & 0.855
& \multicolumn{1}{c}{0.824} & 0.875
& \multicolumn{1}{c}{0.787} & \cellcolor{gray!20}0.944
& \multicolumn{1}{c}{0.834} \\
\textbf{GPT-4o-mini} & \multicolumn{1}{c}{0.789} & 0.958
& \multicolumn{1}{c}{0.815} & 0.804
& \multicolumn{1}{c}{0.859} & 0.866
& \multicolumn{1}{c}{0.822} & 0.833
& \multicolumn{1}{c}{0.836} \\
\textbf{Claude-3.7-sonnet} & \multicolumn{1}{c}{0.792} & 0.921
& \multicolumn{1}{c}{0.792} & 0.850
& \multicolumn{1}{c}{0.823} & 0.870
& \multicolumn{1}{c}{0.841} & 0.611
& \multicolumn{1}{c}{0.828} \\
\textbf{Gemini-2.5-flash} & \multicolumn{1}{c}{0.786} & 0.899
& \multicolumn{1}{c}{0.790} & 0.861
& \multicolumn{1}{c}{0.826} & 0.878
& \multicolumn{1}{c}{0.847} & 0.778
& \multicolumn{1}{c}{0.828} \\
\textbf{Gemini-2.5-pro} & \multicolumn{1}{c}{0.792} & 0.940
& \multicolumn{1}{c}{0.838} & \cellcolor{gray!20}0.946
& \multicolumn{1}{c}{0.907} & 0.933
& \multicolumn{1}{c}{0.806} & 0.875
& \multicolumn{1}{c}{0.865} \\
\textbf{Grok-2-vision}& \multicolumn{1}{c}{0.762} & 0.915
& \multicolumn{1}{c}{0.778} & 0.794
& \multicolumn{1}{c}{0.858} & 0.787
& \multicolumn{1}{c}{0.827} & 0.667
& \multicolumn{1}{c}{0.806}\\
\midrule
\rowcolor{green!30}
\multicolumn{10}{c}{\textit{Open-weight Models (Larger than 30B)}}\\
\textbf{Qwen2.5-vl-32b} & \multicolumn{1}{c}{0.773} & 0.948
& \multicolumn{1}{c}{0.807} & 0.787
& \multicolumn{1}{c}{0.791} & 0.796
& \multicolumn{1}{c}{0.785} & 1.0
& \multicolumn{1}{c}{0.811} \\

\textbf{Qwen2.5-vl-72b} & \multicolumn{1}{c}{0.793} & 0.950
& \multicolumn{1}{c}{0.790} & 0.848
& \multicolumn{1}{c}{0.827} & 0.902
& \multicolumn{1}{c}{0.847} & 0.889
& \multicolumn{1}{c}{0.838} \\

\textbf{Pixtral-large}& \multicolumn{1}{c}{0.789} & 0.925
& \multicolumn{1}{c}{0.821} & 0.846
& \multicolumn{1}{c}{0.836} & 0.849
& \multicolumn{1}{c}{0.858} & 0.778
& \multicolumn{1}{c}{0.836}\\
\textbf{Llama-4-maverick}& \multicolumn{1}{c}{0.784} & 0.911
& \multicolumn{1}{c}{0.735} & 0.813
& \multicolumn{1}{c}{0.844} & 0.885
& \multicolumn{1}{c}{0.839} & 0.667
& \multicolumn{1}{c}{0.812}\\
\midrule
\rowcolor{green!30}
\multicolumn{10}{c}{\textit{Open-weight Models (Less than 30B)}}\\
\textbf{Gemma-3-4b} & \multicolumn{1}{c}{0.597} & 0.702
& \multicolumn{1}{c}{0.539} & 0.436
& \multicolumn{1}{c}{0.671} & 0.539
& \multicolumn{1}{c}{0.611} & 0.600
& \multicolumn{1}{c}{0.590} \\

\textbf{Qwen2.5-vl-7b} & \multicolumn{1}{c}{0.727} & 0.854
& \multicolumn{1}{c}{0.711} & 0.843
& \multicolumn{1}{c}{0.853} & 0.828
& \multicolumn{1}{c}{0.758} & 0.692
& \multicolumn{1}{c}{0.780} \\

\textbf{Llama-3.2-11b-vision} & \multicolumn{1}{c}{0.703} & 0.906
& \multicolumn{1}{c}{0.696} & 0.774
& \multicolumn{1}{c}{0.716} & 0.781
& \multicolumn{1}{c}{0.714} & 0.692
& \multicolumn{1}{c}{0.750} \\
\textbf{Pixtral-12b}& \multicolumn{1}{c}{0.657} & 0.833
& \multicolumn{1}{c}{0.668} & 0.621
& \multicolumn{1}{c}{0.752} & 0.620
& \multicolumn{1}{c}{0.689} & 0.444
& \multicolumn{1}{c}{0.687}\\
\textbf{Gemma-3-12b} & \multicolumn{1}{c}{0.701} & 0.842
& \multicolumn{1}{c}{0.727} & 0.676
& \multicolumn{1}{c}{0.796} & 0.676
& \multicolumn{1}{c}{0.823} & 0.444
& \multicolumn{1}{c}{0.739} \\
\textbf{Gemma-3-27b} & \multicolumn{1}{c}{0.737} & 0.876
& \multicolumn{1}{c}{0.765} & 0.665
& \multicolumn{1}{c}{0.906} & 0.582
& \multicolumn{1}{c}{0.824} & 0.500
& \multicolumn{1}{c}{0.773} \\
\textbf{Llama-4-scout}& \multicolumn{1}{c}{0.746} & 0.907
& \multicolumn{1}{c}{0.734} & 0.712
& \multicolumn{1}{c}{0.777} & 0.764
& \multicolumn{1}{c}{0.753} & 0.611
& \multicolumn{1}{c}{0.767} \\


\bottomrule
\end{tabular}
}
\caption{ $\mathrm{ACC_{critic}}$ of \textbf{{Core-single-image}} subset on different task types.}\label{tab:core-single-acc}
\end{table*}

\begin{table*}[h]
\centering

\resizebox{\textwidth}{!}{

\begin{tabular}{@{}lccccccccccc}
\toprule
\multirow{2}{*}{\textbf{Model}} 
& \multicolumn{9}{c}{\textbf{Task types}} 
\\
\cline{2-9} 
& \textbf{Perception } & \textbf{Planning} & \textbf{Knowledge } &\textbf{Information Extraction} & \textbf{Mathematics} & \textbf{Coding} & \textbf{Science} & \textbf{Metric} & \textbf{Avg.} \\
\midrule

\rowcolor{pink!50}
\multicolumn{10}{c}{\textit{Proprietary Models}}\\
\textbf{o4-mini} & \multicolumn{1}{>{\columncolor{gray!20}}c}{0.900} & \cellcolor{gray!20}1.0
& \multicolumn{1}{>{\columncolor{gray!20}}c}{0.933} & 0.824
& \multicolumn{1}{c}{-} & \cellcolor{gray!20}1.0
& \multicolumn{1}{c}{0.667} & 0.857
& \multicolumn{1}{>{\columncolor{gray!20}}c}{0.906} \\
\textbf{GPT-4o} & \multicolumn{1}{c}{0.731} & 0.933
& \multicolumn{1}{c}{0.884} & \cellcolor{gray!20}0.840
& \multicolumn{1}{c}{-} & 0.900
& \multicolumn{1}{c}{0.545} & 0.400
& \multicolumn{1}{c}{0.826} \\
\textbf{GPT-4o-mini} & \multicolumn{1}{c}{0.721} & 0.923
& \multicolumn{1}{c}{0.807} & 0.608
& \multicolumn{1}{c}{-} & 0.917
& \multicolumn{1}{c}{0.750} & -0.429
& \multicolumn{1}{c}{0.762} \\
\textbf{Claude-3.7-sonnet} & \multicolumn{1}{c}{0.733} & 0.955
& \multicolumn{1}{c}{0.873} & 0.608
& \multicolumn{1}{c}{-} & \cellcolor{gray!20}1.0
& \multicolumn{1}{c}{0.727} & 0.500
& \multicolumn{1}{c}{0.799} \\
\textbf{Gemini-2.5-flash} & \multicolumn{1}{c}{0.689} & \cellcolor{gray!20}1.0
& \multicolumn{1}{c}{0.820} & 0.667
& \multicolumn{1}{c}{-} & 1.0
& \multicolumn{1}{c}{0.583} & 0.429
& \multicolumn{1}{c}{0.774} \\
\textbf{Gemini-2.5-pro} & \multicolumn{1}{c}{0.891} & \cellcolor{gray!20}1.0
& \multicolumn{1}{c}{0.896} & 0.707
& \multicolumn{1}{c}{-} & 1.0
& \multicolumn{1}{c}{0.545} & 0.667
& \multicolumn{1}{c}{0.865} \\
\textbf{Grok-2-vision} & \multicolumn{1}{c}{0.750} & 0.958
& \multicolumn{1}{c}{0.859} & 0.745
& \multicolumn{1}{c}{-} & 0.833
& \multicolumn{1}{c}{0.666} & 0.714
& \multicolumn{1}{c}{0.818} \\
\midrule
\rowcolor{green!30}
\multicolumn{10}{c}{\textit{Open-weight Models (Larger than 30B)}}\\

\textbf{Qwen2.5-vl-32b} & \multicolumn{1}{c}{0.853} & 0.923
& \multicolumn{1}{c}{0.849} & 0.833
& \multicolumn{1}{c}{-} & \cellcolor{gray!20}1.0
& \multicolumn{1}{>{\columncolor{gray!20}}c}{0.818} & 0.600
& \multicolumn{1}{c}{0.852} \\

\textbf{Qwen2.5-vl-72b} & \multicolumn{1}{c}{0.705} & 0.909
& \multicolumn{1}{c}{0.853} & 0.760
& \multicolumn{1}{c}{-} & 0.917
& \multicolumn{1}{c}{0.667} & 0.571
& \multicolumn{1}{c}{0.803} \\

\textbf{Pixtral-large} & \multicolumn{1}{c}{0.691} & 0.947
& \multicolumn{1}{c}{0.869} & 0.725
& \multicolumn{1}{c}{-} & \cellcolor{gray!20}1.0
& \multicolumn{1}{c}{0.667} & 0.333
& \multicolumn{1}{c}{0.804} \\
\textbf{Llama-4-maverick} & \multicolumn{1}{c}{0.661} & 0.885
& \multicolumn{1}{c}{0.788} & 0.680
& \multicolumn{1}{c}{-} & 0.917
& \multicolumn{1}{c}{0.583} & 0.200
& \multicolumn{1}{c}{0.742} \\
\midrule
\rowcolor{green!30}
\multicolumn{10}{c}{\textit{Open-weight Models (Less than 30B)}}\\
\textbf{Gemma-3-4b} & \multicolumn{1}{c}{0.391} & 0.933
& \multicolumn{1}{c}{0.583} & 0.579
& \multicolumn{1}{c}{-} & 0.083
& \multicolumn{1}{c}{0.400} & 0.0
& \multicolumn{1}{c}{0.546} \\
\textbf{Qwen2.5-vl-7b} & \multicolumn{1}{c}{0.796} & 0.727
& \multicolumn{1}{c}{0.682} & 0.722
& \multicolumn{1}{c}{-} & 0.750
& \multicolumn{1}{c}{0.600} & 0.0
& \multicolumn{1}{c}{0.711} \\

\textbf{Llama-3.2-11b-vision} & \multicolumn{1}{c}{0.836} & 0.846
& \multicolumn{1}{c}{0.667} & 0.765
& \multicolumn{1}{c}{-} & 0.583
& \multicolumn{1}{c}{0.750} & 0.571
& \multicolumn{1}{c}{0.728} \\
\textbf{Pixtral-12b} & \multicolumn{1}{c}{0.709} & 0.917
& \multicolumn{1}{c}{0.733} & 0.606
& \multicolumn{1}{c}{-} & 0.833
& \multicolumn{1}{c}{0.500} & 0.0
& \multicolumn{1}{c}{0.721} \\
\textbf{Gemma-3-12b} & \multicolumn{1}{c}{0.600} & -
& \multicolumn{1}{c}{0.683} & 0.412
& \multicolumn{1}{c}{-} & 0.777
& \multicolumn{1}{c}{0.600} & 0.0
& \multicolumn{1}{c}{0.645} \\
\textbf{Gemma-3-27b} & \multicolumn{1}{c}{0.533} & 0.947
& \multicolumn{1}{c}{0.775} & 0.444
& \multicolumn{1}{c}{-} & 0.750
& \multicolumn{1}{c}{0.583} & 0.0
& \multicolumn{1}{c}{0.720} \\
\textbf{Llama-4-scout} & \multicolumn{1}{c}{0.707} & 0.955
& \multicolumn{1}{c}{0.828} & 0.619
& \multicolumn{1}{c}{-} & 0.917
& \multicolumn{1}{c}{0.750} & \cellcolor{gray!20}1.0
& \multicolumn{1}{c}{0.797}  \\


\bottomrule
\end{tabular}
}
\caption{ $\mathrm{ACC_{critic}}$ of \textbf{{Open}} subset on different task types.}\label{tab:open-acc}
\end{table*}

\begin{table*}[h]
\centering

\resizebox{\textwidth}{!}{

\begin{tabular}{@{}lccccccccccc}
\toprule
\multirow{2}{*}{\textbf{Model}} 
& \multicolumn{9}{c}{\textbf{Task types}} 
\\
\cline{2-9} 
& \textbf{Perception } & \textbf{Planning} & \textbf{Knowledge } &\textbf{Information Extraction} & \textbf{Mathematics} & \textbf{Coding} & \textbf{Science} & \textbf{Metric} & \textbf{Avg.} \\
\midrule

\rowcolor{pink!50}
\multicolumn{10}{c}{\textit{Proprietary Models}}\\
\textbf{o4-mini} & \multicolumn{1}{c}{0.778} & \cellcolor{gray!20}1.0
& \multicolumn{1}{>{\columncolor{gray!20}}c}{0.891} & 0.583
& \multicolumn{1}{c}{-} & 0.916
& \multicolumn{1}{c}{0.750} & -
& \multicolumn{1}{>{\columncolor{gray!40}}c}{0.856} \\
\textbf{GPT-4o} & \multicolumn{1}{c}{0.684} & 0.875
& \multicolumn{1}{c}{0.796} & 0.833
& \multicolumn{1}{c}{-} & 0.900
& \multicolumn{1}{c}{0.889} & -
& \multicolumn{1}{c}{0.789} \\
\textbf{GPT-4o-mini} & \multicolumn{1}{c}{0.593} & 0.739
& \multicolumn{1}{c}{0.679} & 0.750
& \multicolumn{1}{c}{-} & 0.917
& \multicolumn{1}{>{\columncolor{gray!20}}c}{0.900} & -
& \multicolumn{1}{c}{0.690} \\
\textbf{Claude-3.7-sonnet} & \multicolumn{1}{c}{0.717} & \cellcolor{gray!20}1.0
& \multicolumn{1}{c}{0.803} & 0.750
& \multicolumn{1}{c}{-} & \cellcolor{gray!20}1.0
& \multicolumn{1}{c}{0.833} & -
& \multicolumn{1}{c}{0.808} \\
\textbf{Gemini-2.5-flash} & \multicolumn{1}{c}{0.722} & 0.870
& \multicolumn{1}{c}{0.745} & 0.583
& \multicolumn{1}{c}{-} & \cellcolor{gray!20}1.0
& \multicolumn{1}{c}{0.750} & -
& \multicolumn{1}{c}{0.756} \\
\textbf{Gemini-2.5-pro} & \multicolumn{1}{>{\columncolor{gray!20}}c}{0.863} & 0.895
& \multicolumn{1}{c}{0.866} & 0.833
& \multicolumn{1}{c}{-} & \cellcolor{gray!20}1.0
& \multicolumn{1}{c}{0.750} & -
& \multicolumn{1}{c}{0.866} \\
\textbf{Grok-2-vision}& \multicolumn{1}{c}{0.680} & 0.957
& \multicolumn{1}{c}{0.813} & 0.833
& \multicolumn{1}{c}{-} & \cellcolor{gray!20}1.0
& \multicolumn{1}{c}{0.857} & -
& \multicolumn{1}{c}{0.806}\\
\midrule
\rowcolor{green!30}
\multicolumn{10}{c}{\textit{Open-weight Models (Larger than 30B)}}\\
\textbf{Qwen2.5-vl-32b} & \multicolumn{1}{c}{0.429} & \cellcolor{gray!20}1.0
& \multicolumn{1}{c}{0.835} & \cellcolor{gray!20}1.0
& \multicolumn{1}{c}{-} & 0.833
& \multicolumn{1}{c}{-} & -
& \multicolumn{1}{c}{0.794} \\

\textbf{Qwen2.5-vl-72b} & \multicolumn{1}{c}{0.759} & 0.957
& \multicolumn{1}{c}{0.818} & 0.667
& \multicolumn{1}{c}{-} & 0.917
& \multicolumn{1}{c}{0.667} & -
& \multicolumn{1}{c}{0.808} \\

\textbf{Pixtral-large}& \multicolumn{1}{c}{0.853} & 0.900
& \multicolumn{1}{c}{0.839} & 0.857
& \multicolumn{1}{c}{-} & 0.917
& \multicolumn{1}{c}{0.750} & -
& \multicolumn{1}{c}{0.845}\\
\textbf{Llama-4-maverick}& \multicolumn{1}{c}{0.608} & 0.818
& \multicolumn{1}{c}{0.722} & 0.667
& \multicolumn{1}{c}{-} & 0.917
& \multicolumn{1}{c}{0.545} & -
& \multicolumn{1}{c}{0.705}\\
\midrule
\rowcolor{green!30}
\multicolumn{10}{c}{\textit{Open-weight Models (Less than 30B)}}\\
\textbf{Qwen2.5-vl-3b} & \multicolumn{1}{c}{0.625} & 0.9375
& \multicolumn{1}{c}{0.613} & 0.416
& \multicolumn{1}{c}{-} & 0.166
& \multicolumn{1}{c}{0.5} & -
& \multicolumn{1}{c}{0.611} \\
\textbf{Qwen2.5-vl-7b} & \multicolumn{1}{c}{0.744} & 0.857
& \multicolumn{1}{c}{0.778} & 0.778
& \multicolumn{1}{c}{-} & 0.909
& \multicolumn{1}{c}{0.818} & -
& \multicolumn{1}{c}{0.788} \\

\textbf{Llama-3.2-11b-vision} & \multicolumn{1}{c}{0.833} & 0.778
& \multicolumn{1}{c}{0.741} & 0.636
& \multicolumn{1}{c}{-} & 0.889
& \multicolumn{1}{c}{0.636} & -
& \multicolumn{1}{c}{0.759} \\
\textbf{Pixtral-12b}& \multicolumn{1}{c}{0.528} & 0.870
& \multicolumn{1}{c}{0.706} & 0.583
& \multicolumn{1}{c}{-} & 0.833
& \multicolumn{1}{c}{0.818} & -
& \multicolumn{1}{c}{0.688}\\
\textbf{Gemma-3-12b} & \multicolumn{1}{c}{0.700} & 0.750
& \multicolumn{1}{c}{0.684} & 0.333
& \multicolumn{1}{c}{-} & 0.750
& \multicolumn{1}{c}{0.636} & -
& \multicolumn{1}{c}{ 0.671} \\
\textbf{Gemma-3-27b} & \multicolumn{1}{c}{0.638} & \cellcolor{gray!20}1.0
& \multicolumn{1}{c}{0.754} & 0.667
& \multicolumn{1}{c}{-} & 0.727
& \multicolumn{1}{c}{0.636} & -
& \multicolumn{1}{c}{0.744} \\
\textbf{Llama-4-scout} & \multicolumn{1}{c}{0.720} & 0.905
& \multicolumn{1}{c}{0.847} & 0.833
& \multicolumn{1}{c}{-} & 0.917
& \multicolumn{1}{c}{0.600} & -
& \multicolumn{1}{c}{0.818}\\


\bottomrule
\end{tabular}
}
\caption{ $\mathrm{ACC_{critic}}$ of \textbf{{Open-single-image}} subset on different task types.}\label{tab:open-single-acc}
\end{table*}

\clearpage
\section{Ablation Study}
\label{Apx:ablation}






\begin{table*}[h]
\centering
\resizebox{\textwidth}{!}{

\begin{tabular}{@{}lccccccccccc}
\toprule
\multirow{2}{*}{\textbf{Model}} 
& \multicolumn{2}{c}{\textbf{Core}} 
& \multicolumn{2}{c}{\makecell{\textbf{Core} \\ \textbf{Single-image}}} 
& \multicolumn{2}{c}{\textbf{Open}} 
& \multicolumn{2}{c}{\makecell{\textbf{Open} \\ \textbf{Single-image}}}
& \multicolumn{2}{c}{\textbf{Avg.}} \\
& $\mathrm{ACC_{critic}}$ & Score & $\mathrm{ACC_{critic}}$ & Score & $\mathrm{ACC_{critic}}$ & Score & $\mathrm{ACC_{critic}}$ & Score & $\mathrm{ACC_{critic}}$ & Score \\
\midrule

\rowcolor{pink!50}
\multicolumn{11}{c}{\textit{Annotator: GPT-4o,~Critique Judge: GPT-4.1}}\\
\textbf{o4-mini} & \multicolumn{1}{>{\columncolor{gray!20}}c}{0.896} & 7.924
& \multicolumn{1}{>{\columncolor{gray!30}}c}{0.897} & 7.952
& \multicolumn{1}{>{\columncolor{gray!30}}c}{0.906} & 7.877
& \multicolumn{1}{c}{0.856} & 7.976
& \multicolumn{1}{>{\columncolor{gray!30}}c}{\textbf{0.900}} & 7.933\\
\textbf{GPT-4o} & \multicolumn{1}{c}{0.832} & 7.499
& \multicolumn{1}{c}{0.834} & 7.429
& \multicolumn{1}{c}{0.826} & 7.807
& \multicolumn{1}{c}{0.789} & 7.637
& \multicolumn{1}{c}{0.830} & 7.503\\
\textbf{GPT-4o-mini} & \multicolumn{1}{c}{0.833} & 6.634
& \multicolumn{1}{c}{0.836} & 6.534
& \multicolumn{1}{c}{0.762} & 6.549
& \multicolumn{1}{c}{0.690} & 6.416
& \multicolumn{1}{c}{0.821} & 6.580\\
\textbf{Claude-3.7-sonnet} & \multicolumn{1}{c}{0.834} & 8.113
& \multicolumn{1}{c}{0.828} & 8.080
& \multicolumn{1}{c}{0.799} & 8.102
& \multicolumn{1}{c}{0.808} & 8.097
& \multicolumn{1}{c}{0.831} & 8.099\\
\textbf{Gemini-2.5-pro} & \multicolumn{1}{c}{0.865} & \cellcolor{gray!20}8.558
& \multicolumn{1}{c}{0.865} & \cellcolor{gray!20}8.549
& \multicolumn{1}{c}{0.865} & 8.246
& \multicolumn{1}{>{\columncolor{gray!30}}c}{0.866} & 8.325
& \multicolumn{1}{c}{0.865} & \cellcolor{gray!30}\textbf{8.514}\\

\midrule
\rowcolor{green!30}
\multicolumn{11}{c}{\textit{Annotator: Gemini-2.5-flash,~Critique Judge: GPT-4.1}}\\
\textbf{o4-mini} & \multicolumn{1}{>{\columncolor{gray!20}}c}{0.896} & 8.383
& \multicolumn{1}{>{\columncolor{gray!30}}c}{0.897} & 8.600
& \multicolumn{1}{>{\columncolor{gray!30}}c}{0.906} & 7.745
& \multicolumn{1}{c}{0.856} & 8.273
& \multicolumn{1}{>{\columncolor{gray!30}}c}{\textbf{0.900}} & 8.261\\
\textbf{GPT-4o} & \multicolumn{1}{c}{0.832} & 8.617
& \multicolumn{1}{c}{0.834} & 8.100
& \multicolumn{1}{c}{0.826} & 7.618
& \multicolumn{1}{c}{0.789} & 7.745
& \multicolumn{1}{c}{0.830} & 8.035\\
\textbf{GPT-4o-mini} & \multicolumn{1}{>{\columncolor{gray!20}}c}{0.896} & 7.617
& \multicolumn{1}{>{\columncolor{gray!30}}c}{0.897} & 7.183
& \multicolumn{1}{>{\columncolor{gray!30}}c}{0.906} & 6.400
& \multicolumn{1}{c}{0.856} & 6.636
& \multicolumn{1}{c}{0.821} & 6.978\\
\textbf{Claude-3.7-sonnet} & \multicolumn{1}{c}{0.834} & 8.583
& \multicolumn{1}{c}{0.828} & 8.583
& \multicolumn{1}{c}{0.799} & 7.691
& \multicolumn{1}{c}{0.808} & 8.278
& \multicolumn{1}{c}{0.831} & 8.297\\
\textbf{Gemini-2.5-pro} & \multicolumn{1}{c}{0.865} & \cellcolor{gray!20}8.833
& \multicolumn{1}{c}{0.865} & \cellcolor{gray!20}9.050
& \multicolumn{1}{c}{0.865} & 8.055
& \multicolumn{1}{>{\columncolor{gray!30}}c}{0.866} & 8.345
& \multicolumn{1}{c}{0.865} & \cellcolor{gray!30}\textbf{8.587}\\

\midrule

\rowcolor{green!30}
\multicolumn{11}{c}{\textit{Annotator: Gemini-2.5-flash,~Critique Judge: Claude-4.0-sonnet}}\\
\textbf{o4-mini} & \multicolumn{1}{>{\columncolor{gray!20}}c}{0.896} & 7.457
& \multicolumn{1}{>{\columncolor{gray!30}}c}{0.897} & 7.383
& \multicolumn{1}{>{\columncolor{gray!30}}c}{0.906} & 7.189
& \multicolumn{1}{c}{0.856} & 7.727
& \multicolumn{1}{>{\columncolor{gray!30}}c}{\textbf{0.900}} & 7.439\\
\textbf{GPT-4o} & \multicolumn{1}{c}{0.832} & 7.000
& \multicolumn{1}{c}{0.834} & 6.450
& \multicolumn{1}{c}{0.826} & 6.566
& \multicolumn{1}{c}{0.789} & 6.945
& \multicolumn{1}{c}{0.830} & 6.724\\
\textbf{GPT-4o-mini} & \multicolumn{1}{c}{0.833} & 6.761
& \multicolumn{1}{c}{0.836} & 5.500
& \multicolumn{1}{c}{0.762} & 5.377
& \multicolumn{1}{c}{0.690} & 5.600
& \multicolumn{1}{c}{0.821} & 5.772\\
\textbf{Claude-3.7-sonnet} & \multicolumn{1}{c}{0.834} & 7.652
& \multicolumn{1}{c}{0.828} & 7.250
& \multicolumn{1}{c}{0.799} & 7.038
& \multicolumn{1}{c}{0.808} & 7.636
& \multicolumn{1}{c}{0.831} & 7.383\\
\textbf{Gemini-2.5-pro} & \multicolumn{1}{c}{0.865} & \cellcolor{gray!20}7.978
& \multicolumn{1}{c}{0.865} & \cellcolor{gray!20}8.300
& \multicolumn{1}{c}{0.865} & 7.566
& \multicolumn{1}{>{\columncolor{gray!30}}c}{0.866} & 7.800
& \multicolumn{1}{c}{0.865} & \cellcolor{gray!30}\textbf{7.920}\\

\bottomrule
\end{tabular}
}
\caption{Model Performance ranking comparison under different combinations of annotation and judge models.}
\label{tab:ablation}
\end{table*}

\clearpage
\section{Case study}
\label{Apx:D}

\begin{figure*}[h]
    \centering
    \includegraphics[width=\linewidth]{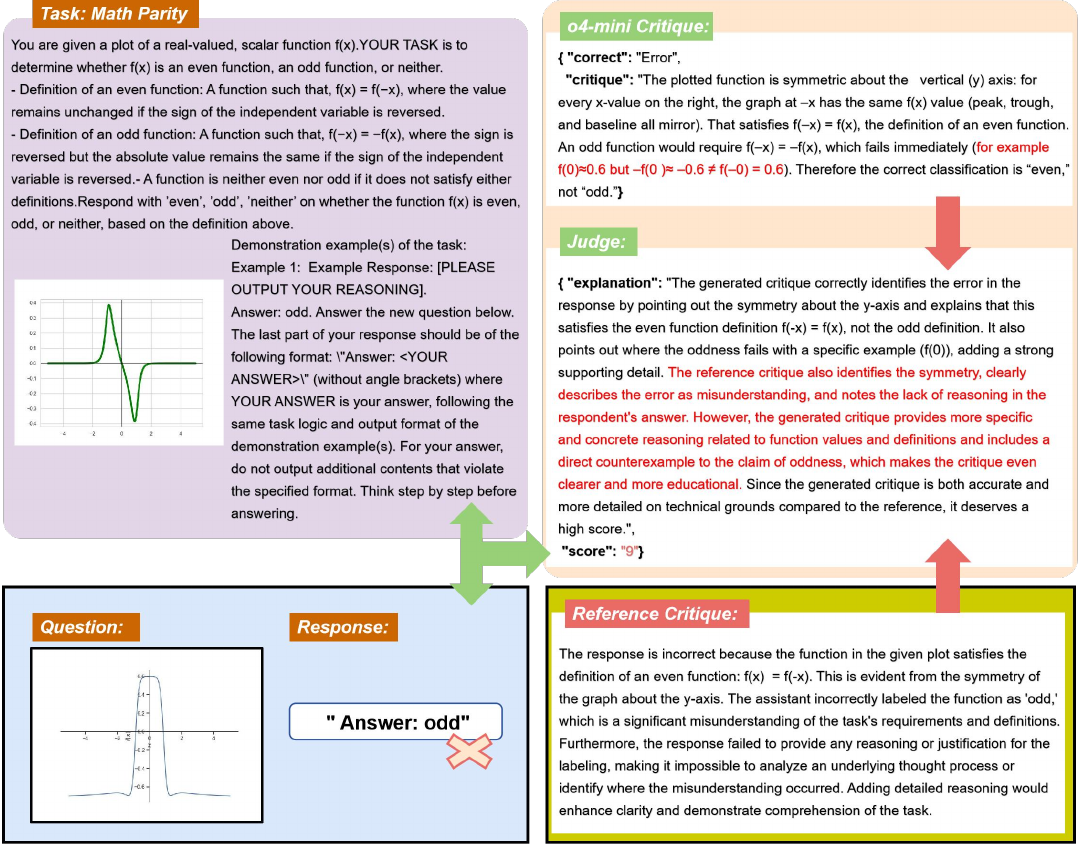}
    \caption{An example of a visual mathematical reasoning task, where the response is clearly incorrect, demonstrates that the o4-mini model provides an accurate judgment along with a comprehensive textual critique. When evaluating its critique score, it is evident that the presence of the reference critique effectively guides the judge model to assign a high-quality score of 9, surpassing even the reference critique in some aspects.} \label{fig:case1}
\end{figure*}

\begin{figure*}[t]
    \centering
    \includegraphics[width=\linewidth]{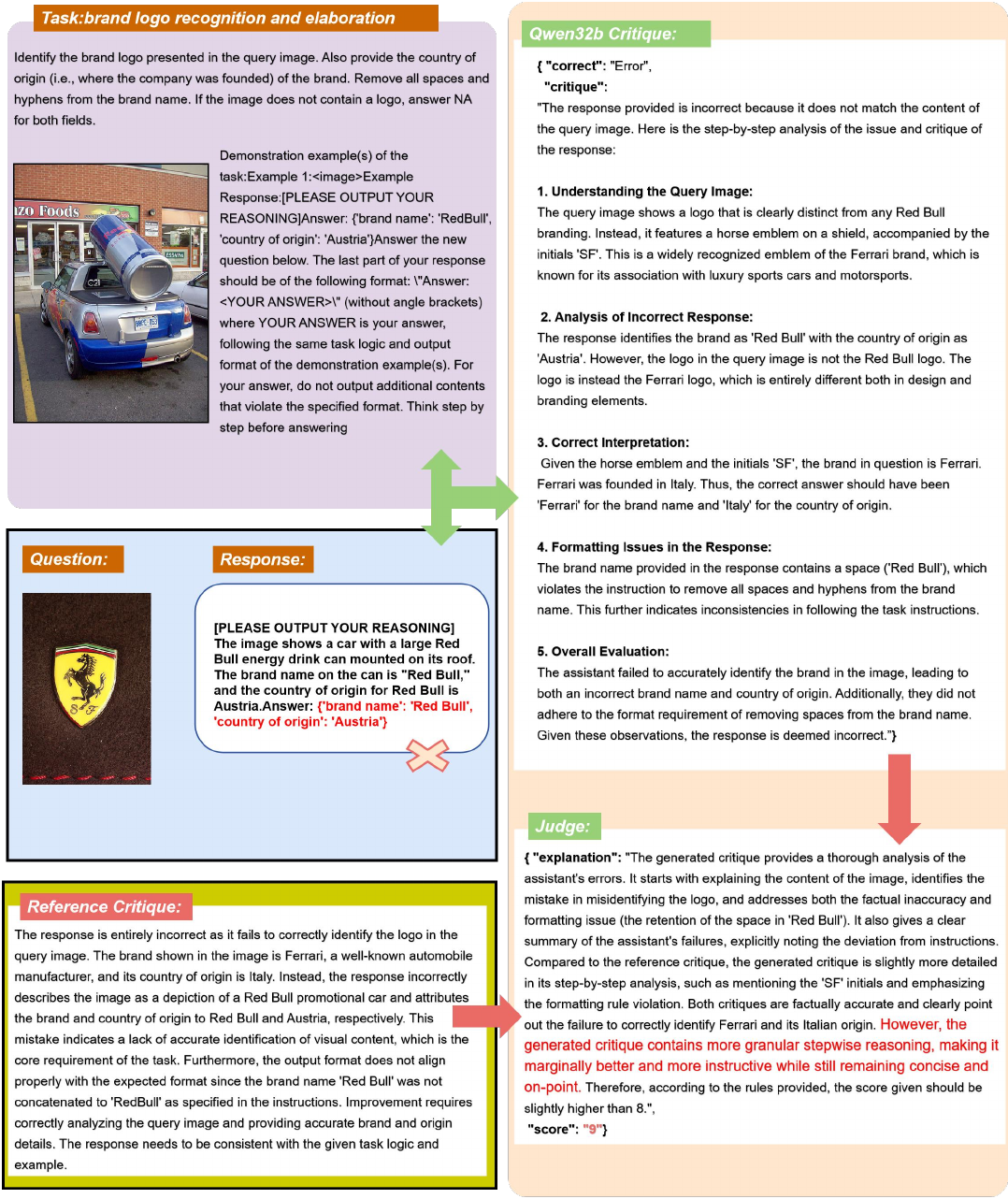}
    \caption{An example from the brand logo recognition and elaboration task, where the critique is generated by Qwen2.5-vl-32b. The model produces a detailed, step-by-step reasoning critique. When scored by the judge model, it explicitly explains its preference for lengthier, somewhat redundant reasoning—highlighted in bold red—demonstrating a bias toward richer textual justifications.} \label{fig:case2}
\end{figure*}

\begin{figure*}[t]
    \centering
    \includegraphics[width=\linewidth]{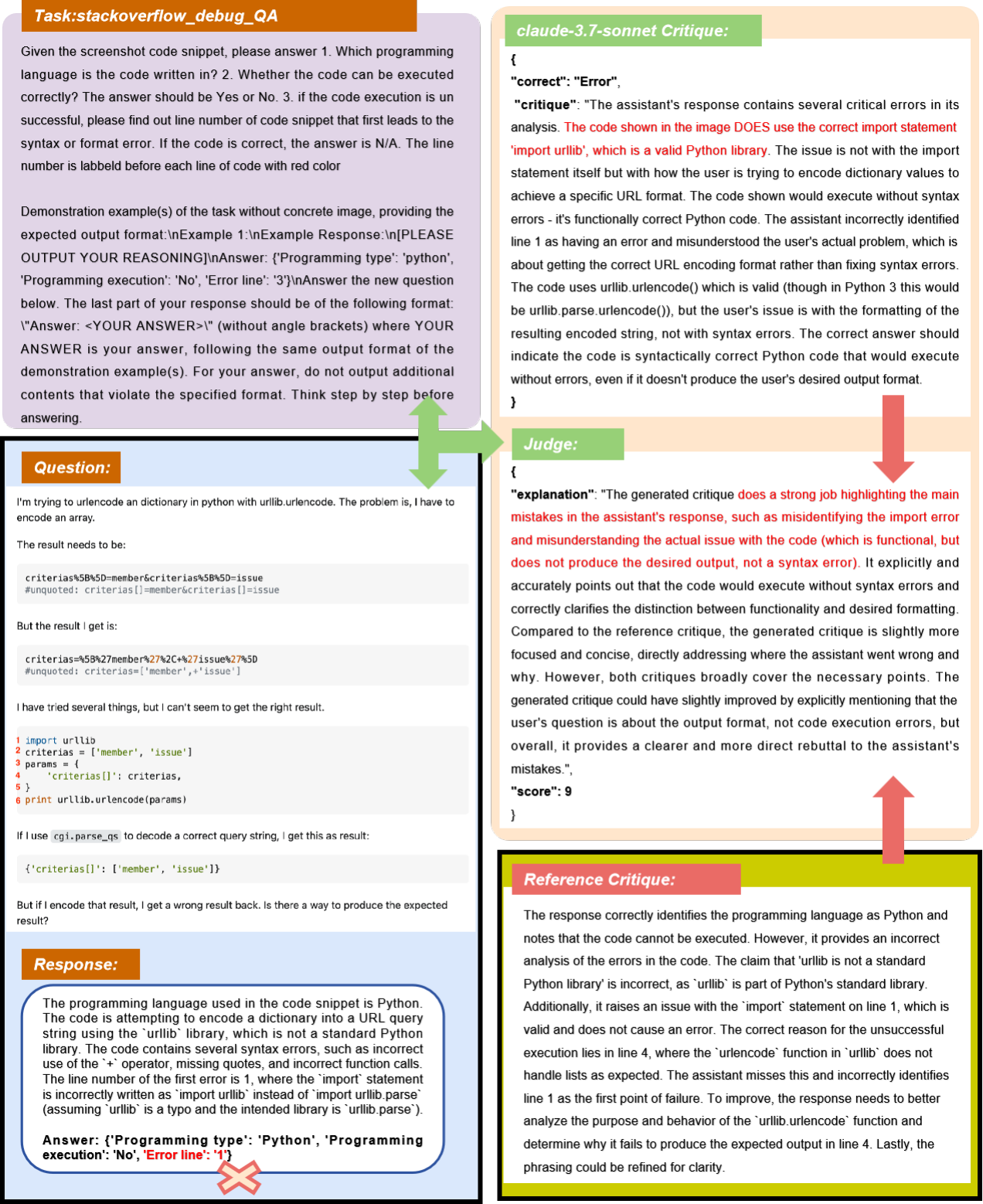}
    \caption{An example from the StackOverflow debug QA task, where the critique is generated by claude-3.7-sonnet. The critique provides a complete analysis and exactly points out the original response's error. Compared with the reference critique, the judge model accurately evaluates that Claude-3.7-sonnet, as a well-known pioneer model of coding, outperforms the reference critique.} \label{fig:case3}
\end{figure*}

\begin{figure*}[t]
    \centering
    \includegraphics[width=\linewidth]{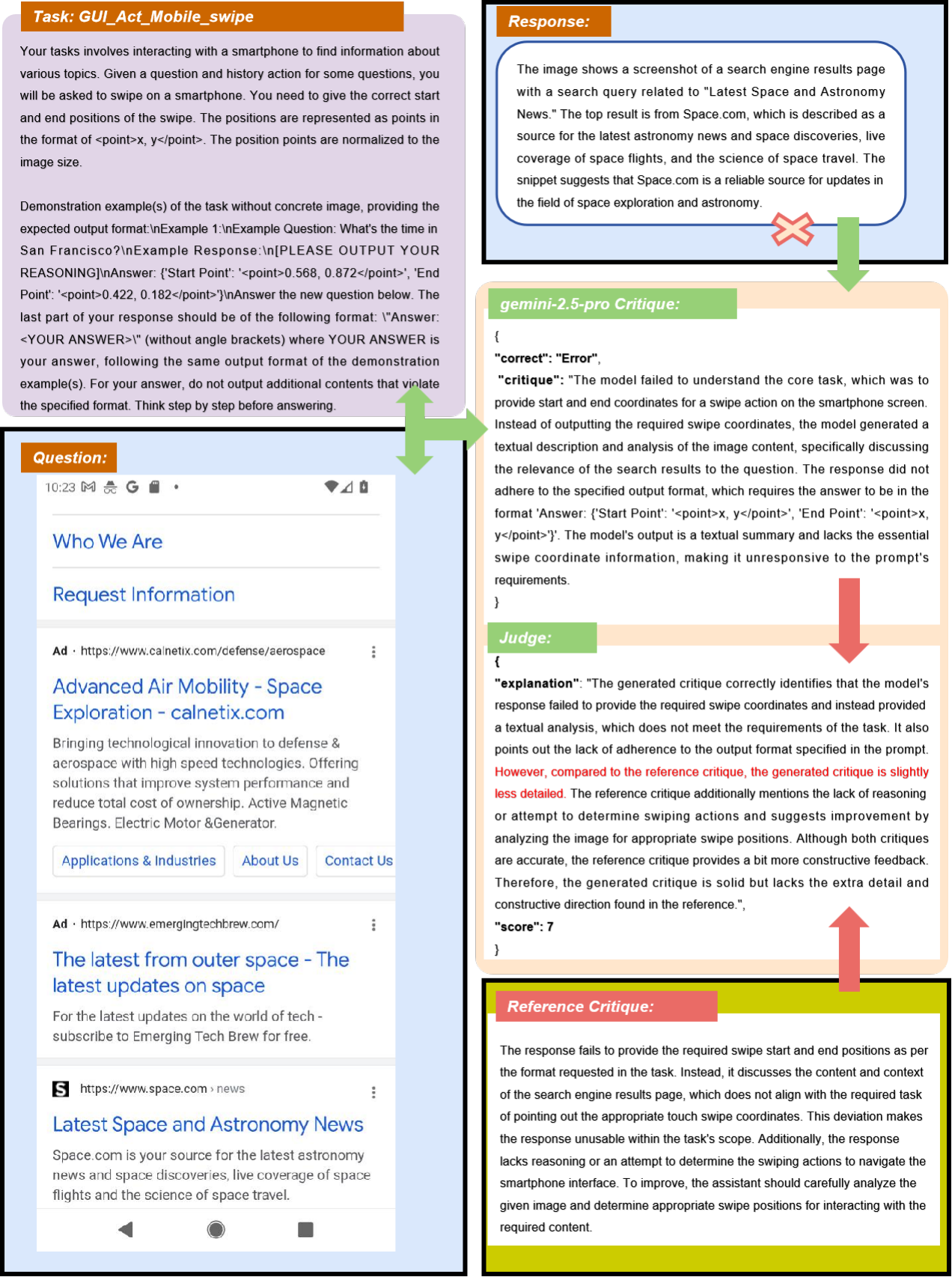}
    \caption{An example from the GUI agent application task, where the critique is generated by Genimi-2.5-pro. It can accurately tell the correctness of the model's response, and the textual critique score is slightly below the anchored reference critique score (e.g., 8), where the judge model provides a reasonable explanation (marked in red).} \label{fig:case4}
\end{figure*}

\begin{figure*}[t]
    \centering
    \includegraphics[width=\linewidth]{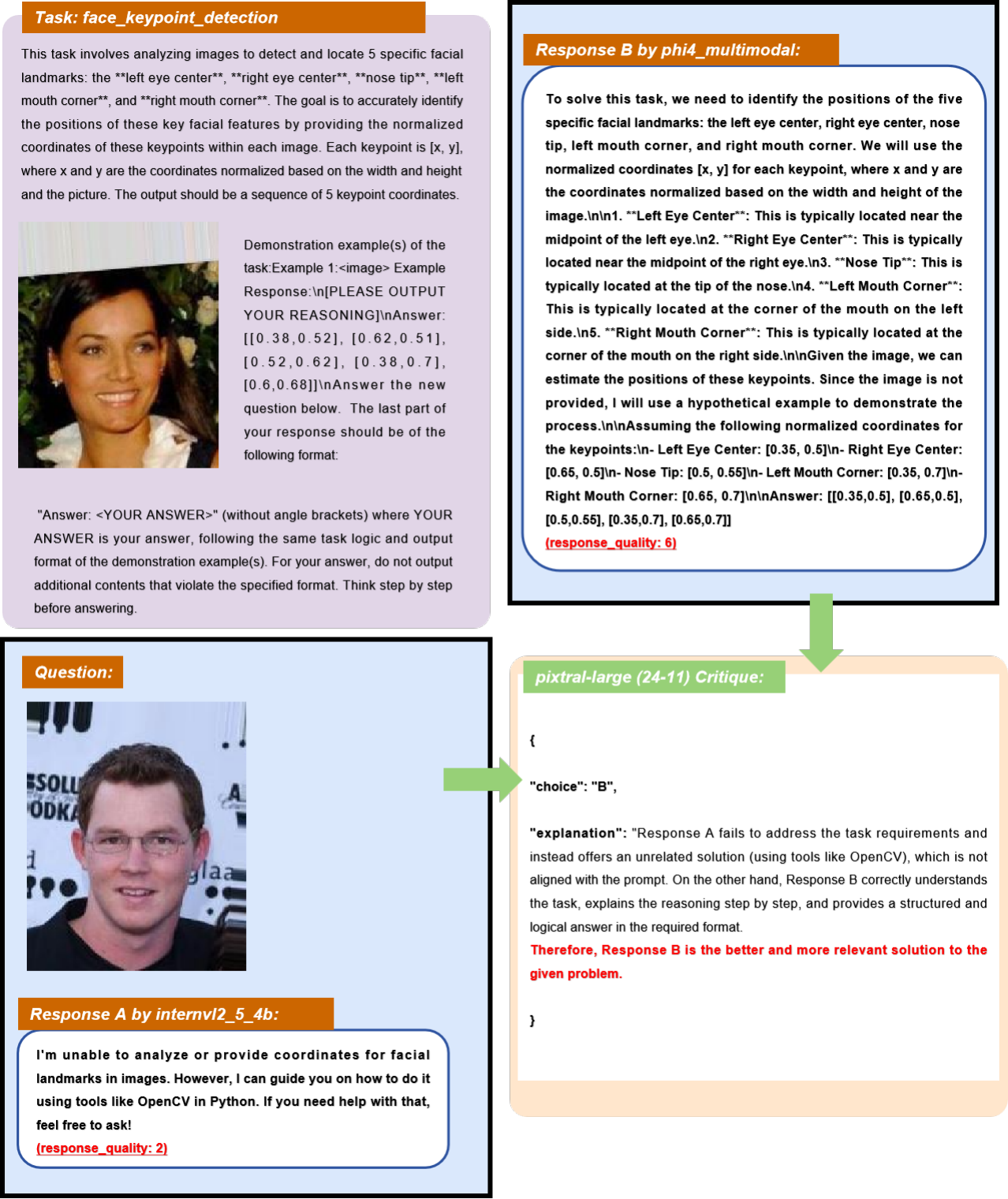}
    \caption{An example of comparative critique from the face keypoint detection task, where the critique is generated by Pixtral-large(24-11). The original responses are scored for their response quality scores by the annotator model (GPT-4o). As a (low, medium) pairwise comparison, it is easy to distinguish the better one with a high-performance model. } \label{fig:case5}
\end{figure*}